\documentclass[journal]{IEEEtran}
\def\@IEEEclspkgerror{\ClassError{IEEEtran}}
\usepackage{enumitem}
\usepackage{cite}
\usepackage{graphicx}
\usepackage{algorithm}
\usepackage{subcaption}
\usepackage{comment}
\usepackage{amsmath}
\usepackage{amssymb}
\usepackage{algpseudocode}
\usepackage{amsthm}

\usepackage{soul}
\usepackage[dvipsnames]{xcolor}
\usepackage{hyperref}
\usepackage[dvipsnames]{xcolor}

\usepackage{array}
\newcolumntype{P}[1]{>{\centering\arraybackslash}p{#1}}

\newcolumntype{M}[1]{>{\centering\arraybackslash}m{#1}}

\makeatletter
\newcounter{parenttheorem}

\makeatother

\def\BibTeX{{\rm B\kern-.05em{\sc i\kern-.025em b}\kern-.08em
    T\kern-.1667em\lower.7ex\hbox{E}\kern-.125emX}}
\usepackage{lineno}

\begin{document}
\title{Learning Position From Vehicle Vibration \\ Using an Inertial Measurement Unit}

\author{Barak Or, \IEEEmembership{Member, IEEE}, Nimrod Segol, Areej Eweida, and, Maxim Freydin, \IEEEmembership{ Member, IEEE}
    \thanks{Submitted on February 2023, revised on March 2023, Accepted on June 2024.}
\thanks{All authors are with ALMA Technologies Ltd, Haifa, 340000, Israel (e-mail: barakorr@gmail.com).}}

\markboth{Learning Position From Vehicle Vibration Using an Inertial Measurement Unit / Or et al}%
{Or et al: Vehicle Positioning Inference by Learning Road Signature From an IMU Sensor}
\maketitle

\begin{abstract}
This paper presents a novel approach to vehicle positioning that operates without reliance on the global navigation satellite system (GNSS). Traditional GNSS approaches are vulnerable to interference in certain environments, rendering them unreliable in situations such as urban canyons, under flyovers, or in low reception areas. This study proposes a vehicle positioning method based on learning the road signature from accelerometer and gyroscope measurements obtained by an inertial measurement unit (IMU) sensor. In our approach, the route is divided into segments, each with a distinct signature that the IMU can detect through the vibrations of a vehicle in response to subtle changes in the road surface. The study presents two different data-driven methods for learning the road segment from IMU measurements. One method is based on convolutional neural networks and the other on ensemble random forest applied to handcrafted features. Additionally, the authors present an algorithm to deduce the position of a vehicle in real-time using the learned road segment. The approach was applied in two positioning tasks: (i) a car along a $6 \ [km]$ route in a dense urban area; (ii) an e-scooter on a $1 \ [km]$ route that combined road and pavement surfaces. 
The mean error between the proposed method's position and the ground truth was approximately $50 \ [m] $  for the car and $30 \ [m]$   for the e-scooter. Compared to a solution based on time integration of the IMU measurements, the proposed approach has a mean error of more than 5 times better for e-scooters and 20 times better for cars. 
\end{abstract}


\section{Introduction}\label{sec:introduction}
\IEEEPARstart{I}{n} various domains and applications, continuous real-time positioning solutions are required. Currently, outdoor positioning technology relies heavily on satellite navigation devices such as GNSS receivers. This includes GPS, GLONASS, Galileo, Beidou, and other satellite navigation systems \cite{langley2017introduction,zangenehnejad2021gnss,farrell2008aided}.  Drivers commonly rely on GNSS technology for positioning and navigation, often using their smartphones for this purpose \cite{zangenehnejad2021gnss,wahlstrom2016imu}.  
However, GNSS requires a line of sight with multiple satellites and is thus unsuitable for indoor positioning applications including navigating through tunnels and indoor parking lots \cite{el2021indoor,francis2020long,al2012indoor,zhao2023data}. Additionally, there are scenarios in which GNSS does not offer sufficient accuracy for certain outdoor positioning applications \cite{zhu2018gnss,jing2022integrity,rizos2013locata}.

Given the limitation of GNSS, there are  approaches that make use of other sensors to compensate. In many scenarios, GNSS measurements are fused with maps and additional sensors to ensure continuous localization and improve accuracy. A key example is autonomous vehicles (AVs) \cite{yeong2021sensor,meng2017robust} where there exists a need for continuous availability beyond that provided by GNSS. It is therefore common to perform  fusion with additional sensors such as cameras, LiDARs, inertial sensors, odometry, and others  \cite{ziebinski2016survey,campbell2018sensor,ibisch2013towards,or2022hybrid}. {Unfortunately, most of these sensors are expensive even at the low-cost version of these products: LiDARs may cost in the hundreds of USD and some cameras in the tens of USD. This contrasts with a consumer grade IMU which costs less than one USD. In addition,} the quality of the data these sensors provide depends on various physical conditions, such as lighting and the presence of occlusions. Hence, there are no accurate and continuous positioning solutions for vehicles that operate 
both indoors and outdoors. This work presents an approach that relies only on a single IMU sensor and is a step towards a continuous solution for a broad range of scenarios and environments.

The positioning approach suggested here applies machine learning to inertial sensor readings. Inertial sensors are commonplace in mobile devices and have a broad range of applications including in mobility and transportation. Indeed, various authors have utilized the IMU sensors for many tasks including identifying driver activity, detecting car accidents, and other \cite{wahlstrom2017smartphone,chan2019comprehensive,mozaffari2020deep, or2024carspeednet}. In the navigation domain, it is common to fuse inertial with GNSS measurements in an INS/GNSS (inertial navigation system) fusion scheme. The IMU typically consists of two sensors: an accelerometer that measures specific force and a gyroscope that measures angular velocity. Positioning is deduced by integrating (with respect to time) both the accelerometer and gyroscope measurements and fusing the high-rate low-accuracy integrated solution with a low-rate high-accuracy GNSS measurement. This approach is sometimes referred to as strap-down inertial navigation (SINS) where it could be said that dead reckoning is performed between each pair of GNSS measurements.  The SINS approach is vulnerable to drift, leading rapidly to large errors in the solution \cite{farrell2008aided}. While this work uses the common and affordable IMU sensor which has the significant advantage of high availability, we do not use dead reckoning. Instead, we use the power of machine learning (ML) and deep learning (DL). Machine learning methods presented impressive results in a wide range of domains in recent years \cite{lecun2015deep,bengio2017deep}. Therefore, the authors of this work propose to utilize them to derive a positioning method from an IMU sensor's {raw measurements} that does not suffer from rapidly increasing drift.

\begin{figure}[ht!]
\centering
\includegraphics[scale=0.142]{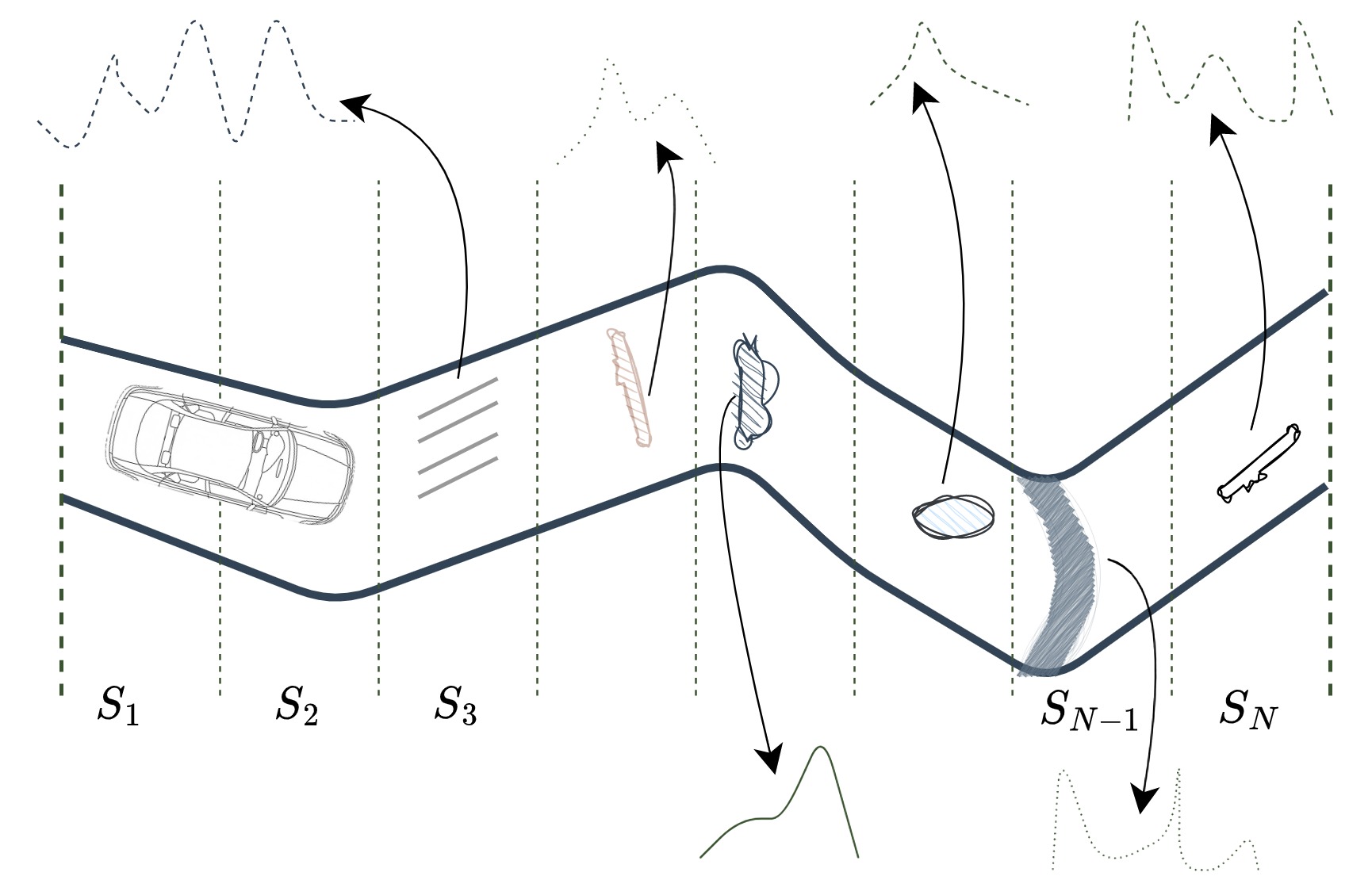}
\caption{An illustration of road segmentation. The proposed methodology is applied for ground vehicle positioning tasks as illustrated. The route is broken into N segments where the IMU measurements are classified into one of the N segments $S_i$, based on unique time-series signal signatures. In this illustration, there is a different road texture in each segment, producing a different and unique IMU signal.}
\label{Fig:seg_illustration}
\end{figure} 

\subsubsection{Related work}
In previous related work, DL-based models were integrated into a wide range of navigation and positioning tasks. In \cite{yan2018ridi} neural networks were trained to regress linear velocities from inertial sensors and use these predictions to constrain accelerometer readings. In   \cite{brossard2019learning}, a DL-based model was developed to address the limitations of wheel odometry in land vehicle navigation.  In \cite{or2022learning}, researchers used long short-term memory (LSTM) networks to learn the road curvature and other features in order to estimate the process noise covariance in a linear Kalman filter \cite{or2021kalman}. In \cite{liu2021vehicle}, an approach to address GNSS outages was proposed involving a DL-based multi-model approach and an extended Kalman filter.  \cite{freydin2022learning}, presents a DL-based model that utilized a stream of measurements from a low-cost IMU to estimate the speed of a car driving in an urban area. In \cite{freydin2022mountnet}, a data-driven approach using a DL-based model is presented to learn the yaw mounting angle of a smartphone equipped with an IMU and strapped to a car. In \cite{shao2018indoor} the authors propose the use of a neural network on WiFi signals for indoor navigation. Previously in \cite{or2023system} the authors introduced a new method to detect road landmarks such as bumps using a machine learning method on IMU sensor readings. {This work significantly extends that past work by deriving a positioning solution} \cite{or2023patent1}. 

{Other related works considered terrain and maps as additional sources of information for solving the positioning problem without GNSS. In }\cite{tianyi2019,adam2008,laftchiev2015,emulapalli2011,eweida2023surface}{, a terrain map based on pitch angles is used to determine position along a previously mapped route. The previous methods are based on IMU and additional car sensors (including wheel speed) as input to Kalman or particle filter to determine position along a known and mapped route. The approaches are challenging to scale and require additional sensors beyond IMUs.}

\subsubsection{Present work}
This study presents a novel approach to vehicle positioning using only IMU measurements, without the need for GNSS or any other sensor. A machine learning model learns the road segment based on vibration measured by the IMU as illustrated in Figure \ref{Fig:seg_illustration} \cite{barak2024system,barak2023system}. This work investigated two different data-driven methods to learn the road segment. The proposed methods were rigorously validated and compared to classical dead reckoning, showing significant improvements in position accuracy for a car along a $6 \ [km]$ urban route and an e-scooter on a $1 \ [km]$ mixed surface route. These findings have important implications for the development of effective and efficient vehicle positioning techniques, particularly in challenging scenarios where traditional approaches may not be feasible. To summarize,
\begin{enumerate}
    \item We propose a positioning method on a given route based only on IMU measurements.
    \item The method is data-driven and we investigated two distinct machine learning approaches that learn from the IMU readings
    \item We collected two data sets, cars, and e-scooters, and validated our approach on these data sets.
    \item We compared our results on these data sets with classical dead reckoning (where accelerometer and gyroscope data is rotated to an inertial frame and integrated with respect to time) and received favorable results.
\end{enumerate}
 
The rest of the paper is organized as follows: Section \ref{sec:method} describes the data-driven approach to vehicle positioning; In particular, we elaborate on the data processing and data-driven ML and DL algorithms we employ. In Section  \ref{sec:Experiments_data_collection}, a description of the real-life datasets we collected is provided, followed by the experimental results and discussion. Finally, Section \ref{sec:conclusions} presents concludes the work.

\section{Data Driven based Positioning Inference}
\label{sec:method}
 The core concept involves utilizing an ML algorithm to determine a vehicle's current position on a given route. Specifically, a route is divided into N  segments of uniform length and an ML model is trained on IMU data from the {phone's} accelerometer and gyroscope sensors {(fixed to a car)} to infer its current road segment. This technique, where the current road segment is inferred, is referred to as \textit{road segmentation}, and the ML model used for this purpose is called a \textit{road segmentor}. The midpoint of the inferred segment is then reported as the vehicle's position. Refer to Figures \ref{Fig:seg_illustration} and \ref{Fig:system} for an illustration of this methodology and the data flowchart, respectively. {The methods and algorithms described in this work were implemented in Python where PyTorch was used for training and testing DL models, and sklearn was used to train and test ensemble random forest classifiers.}

Section \ref{sec:general_desctiption} provides a detailed description of the positioning algorithm, with Subsections \ref{sec:num_segments}, \ref{sec:processing}, and \ref{sec:classifers} elaborating on the critical components of tuning the number of road segments, processing the data, and designing the ML algorithm, respectively.

\subsection{Algorithm Description}
\label{sec:general_desctiption}

\begin{figure}[ht!]
\centering
\includegraphics[scale=0.07]{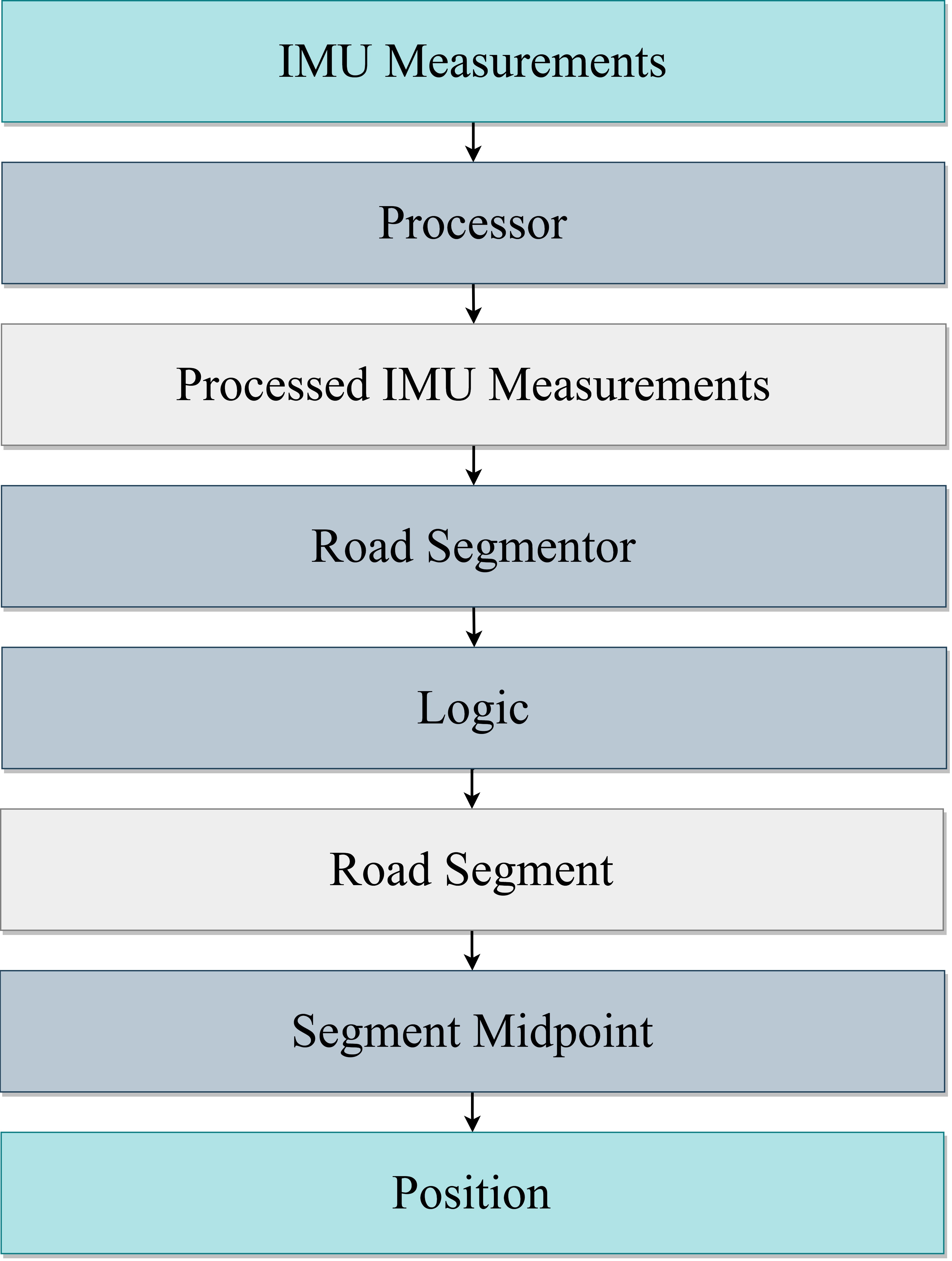}
\caption{Block diagram of the positioning algorithm.}
\label{Fig:system}
\end{figure} 

In this part, a detailed description of the methodology is presented. Initially, input raw measurements from the IMU are obtained, which include acceleration in units of $m/s^{2}$ and angular velocity in units of $rad/s$ across three axes. These measurements are then passed through a processor that is thoroughly described in Subsection \ref{sec:processing}. 
Our data processing approach involves taking $2 \ [s]$ of IMU measurements at a time. Specifically, if the original IMU data is sampled at a frequency of $f$ Hz, then our initial data measurement is represented as $I_t\in \mathbb{R}^{2f \times 6}$ where $t$ represents the time in seconds. The pre-processing stage is denoted by $P$ and is described in detail in the following subsection.
\begin{equation}
     P:\mathbb{R}^{2f\times 6} \to \mathbb{R}^{40\times 6}; \quad  \bar{I}_t = P(I_t).
     \label{eq:preprocessor}
\end{equation}
As part of the pre-processing, we downsample the data to $20$ Hz, resulting in $\bar{I}_t\in \mathbb{R}^{40\times 6}$.
 Next, the processed IMU data is passed through a road segmentor, denoted by $R$, to infer the current road segment. 
 \begin{equation}
 \label{eq:segmentor}
     R:\mathbb{R}^{40\times 6} \to [N]; \quad  \tilde{S}_t = R(\bar{I}_t)
\end{equation}
where $N$ is the number of road segments and $\tilde{S}_t = R(\bar{I}_t)$ is the vehicle's current road segment. 

This study explores two different road segmentors to infer the current road segment from processed IMU data. The first road segmentor is based on a convolutional neural network (CNN) \cite{ismail2019deep}, while the second one employs hand-crafted features (HCF) and an ensemble random forest (Ensemble) \cite{zhang2012ensemble}. Further details can be found in Subsection \ref{sec:classifers}.

The current segment is inferred based on the last segment (which was inferred in the previous time window of two seconds) and the output of the road segmentor. The classifier output is limited to either the current or next segment since the vehicle can only move to the next segment or remain in the current one. Denote this logic by $L$ we have, 
\begin{equation}
\label{eq:logic}
    L:[N] \times [N] \to [N]; \quad  S_t =  L(\tilde{S}_t, S_{t-2})
\end{equation}
Finally, we report the position by taking the midpoints of the inferred segment. That is 
\begin{equation}
    M:[N]\to \mathbb{R}^2;\quad (x_t,y_t) = M(S_t)
\end{equation}
where $M$ is the midpoint of the segment  $S_t$. Based on the described process, a positioning function is provided by
\begin{equation}
  \Phi:\mathbb{R}^{2f\times 6} \times [N] \to  \mathbb{R}^{2};\quad \Phi(I_t, S_{t-2}) =   M\circ L \circ R \circ P
\end{equation}
which takes a raw IMU measurement and the previous inferred segment and outputs a position. The full algorithm is described in a block diagram in Figure \ref{Fig:system} and in Algorithm \ref{ALG:inferposition}.

\begin{figure*}[ht!]
\centering
\includegraphics[scale=0.06]{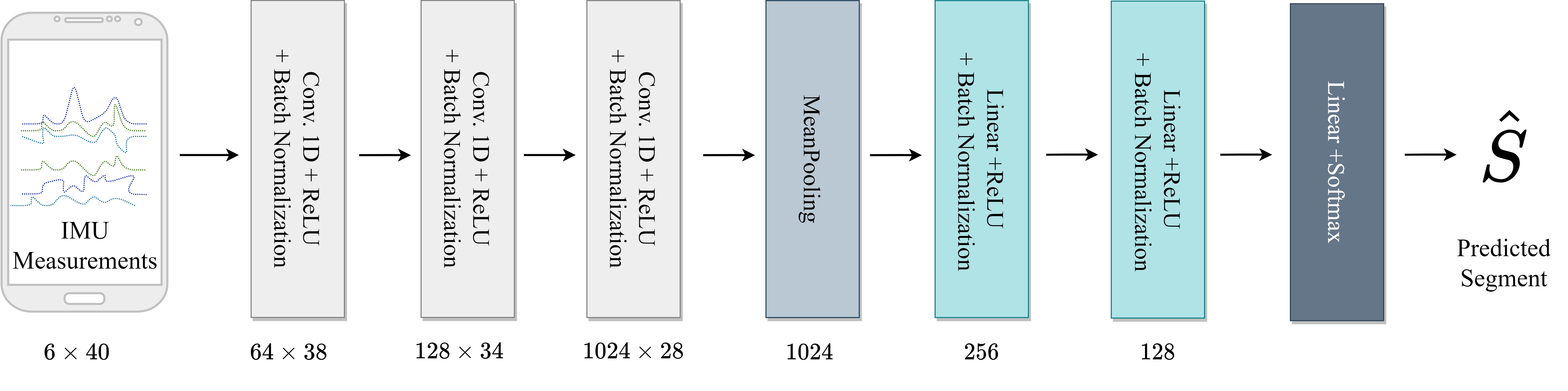}
\caption{The network architecture used for the CNN classifier.}
\label{fig:cnn}
\end{figure*}

\begin{algorithm}
\caption{Positioning from IMU measurements and a predetermined list of segment midpoints}
\label{ALG:inferposition}
\begin{algorithmic}[1]
\State $S_{-2} \gets 1$

\For{$t=0[s], 2[s] \ldots $}
\State $I_t \gets  \text{raw IMU} $ \Comment{Read data from the sensor}
\State $\bar{I}_t \gets  P(I_t) $ \Comment{As described in Subsection \ref{sec:processing}}
\State $\tilde{S}_{t} \gets R(\bar{I}_t)$ \Comment{Road segmentors in Subsection \ref{sec:classifers} }
\If{$\tilde{S}_{t} - S_{t-2} \notin [0, 1]$} 
    \State $S_{t} \gets S_{t-2}$ 
\Else
    \State $S_{t} \gets \tilde{S}_{t}$ 
\EndIf
\State $ (x_t, y_t) \gets M(S_{t})$ \Comment{Report the position as the midpoint of the inferred segment}
\EndFor
\end{algorithmic}
\end{algorithm}

\subsection{Tuning the Number of Segments}\label{sec:num_segments}
The road segmentor takes a processed IMU measurement and outputs a segment number where each of the segments has the same length. Determining the best number of segments in the route is a crucial step in the positioning solution. As the number of segments increases, the road segmentation task becomes more challenging due to the increasing number of features and variations to be captured for each segment. For a given dataset with a limited number of recordings, this also means that there are more segments to distinguish between and fewer samples for each segment to learn from. On the other hand, when the route is divided into smaller segments there is the benefit of increasing the accuracy of distance estimates within each segment. This is because the midpoint of a shorter segment provides a more precise estimate of the vehicle's position within that segment. 

This paper investigates the performance of a positioning solution on two datasets, each with its unique characteristics. We use a data-driven approach to determine the optimal number of road segments for each dataset using a validation set. Figures \ref{fig:num_segs} and \ref{fig:num_segs_matam} present the effect of a number of segments on the average positioning error. We show that the positioning error increases when the number of road segments is either too large or too small. This highlights the importance of choosing an appropriate segment length that is tailored to the specific characteristics of the dataset and the requirements of the application. 

\subsection{Data Processing}
\label{sec:processing}
To prepare the raw IMU data for use by the ML road segmentor, we first apply a series of processing steps to enhance the quality of the data and reduce noise. The raw data used in this study consists of measurements of acceleration in $m/s^2$ and angular velocity in $rad/s$ in three axes. The preprocessor $P$ is applied to raw data as described by Eq. \eqref{eq:preprocessor}. It is composed of,
\begin{enumerate}
\item Low-pass filter (LPF) with a cut-off frequency of 10 Hz. Most IMUs operate at sampling rates at or above 100 Hz which is significantly greater than the dynamics bandwidth of a typical vehicle \cite{rajamani2011vehicle}. This step reduces the noise inherent to low-cost inertial sensors while maintaining key signal information.


\item Down-sample the data to $20$ Hz. To facilitate the use of our ML road segmentor with different mobile phones that sample IMU data at different frequencies, we down-sample the data to a common frequency of 20 Hz. This step is critical for ensuring that our approach is widely applicable and can be used with different types of mobile devices and sensors. 
\item Rotate the measured acceleration and angular velocity vectors to a frame where the third vector component is aligned parallel to the gravity vector (i.e., compensate for the roll and pitch angles \cite{farrell2008aided}). The gravity component is then subtracted from the third component of the acceleration vector. This step makes the data more uniform and independent of the mobile device's roll and pitch installation angles. 

\end{enumerate}

The preprocessing steps outlined above are applied both offline during training of the ML model and online in real-time scenarios. The purpose of these steps is to simplify the raw sensor data by removing noise and other extraneous information that is not relevant to the task of road segmentation and vehicle positioning. It is noted that the above processing procedure has appeared in various prior works e.g. \cite{farrell2008aided, freydin2022learning, freydin2022mountnet}. It is provided for this work to be self-contained.

\subsection{Road Segmentors}
\label{sec:classifers}
In this section, we provide a detailed description of the machine learning methods we employed for road segmentation. Specifically, we introduce the road segmentor $R$, as outlined in Equation \eqref{eq:segmentor}. The road segmentor is a critical component of our approach to vehicle positioning and can be one of two different machine learning methods. We investigate two common data-driven methods to learn position from vibrations, 

\subsubsection{Convolutional neural network}
The first road segmentor  we used is a CNN composed of the following layers:
\begin{itemize}
\item {\bf Linear layer}: A linear transformation is applied to the data from the preceding layer. 
\item {\bf Conv1D layer}: A convolutional, one-dimensional layer creates a convolution kernel that is convolved with the layer input over a single dimension to produce a vector of outputs. 
The dimensions of the convolutions layers  are $64$, $128$, and $1024$ with kernel sizes $3$, $5$ and $7$.
\item {\bf Mean pooling}: Pooling layers help with better generalization capability as they perform a down-sampling feature map.
\item {\bf ReLU}: A nonlinear activation function defined by ReLU$(\alpha)= \max(\alpha, 0)$.
\item {\bf Batch normalization}:  Batch normalization reduces covariate shift. The batch normalization was added after every Conv1D layer (together with the ReLU layer). 
\end{itemize}
See Figure \ref{fig:cnn} for a schematic of the network. The CNN was trained for a total of 200 epochs using the cross-entropy loss function, which is commonly used in classification tasks. Specifically, we define the output of the CNN as $\tilde{y}(I, {\bf W})$, where $\bf W$ represents the weights of the CNN, and $I$ is the preprocessed IMU input. The cross-entropy loss function is then defined as:
\begin{equation}
\ell  =  - \sum\limits_{c = 1}^M {{y_c}} \log \left( {\tilde y\left( {I,{\bf{W}}} \right)} \right)
\end{equation}
where  $y_c= 1$ if the true label $y$ is equal to $c$ and $0$ otherwise. 
We used the ADAM optimizer \cite{kingma2014adam} with an initial learning rate of $0.001$ and multiplicative learning rate decay of $0.1$ every $50$ epochs. We trained
the model using a Tesla V100 Nvidia GPU which took about 24 minutes per epoch. {The run time of the model was evaluated on an ASUS PC with Windows 10 and Intel Core i7 processor achieving a 2 milliseconds per model evaluation. }

\subsubsection{Ensemble random forest}
We evaluated a second approach for road segmentation, the ensemble random forest method. This method takes as input $702$ hand-crafted features extracted from the pre-processed IMU measurements. The feature extraction process is detailed in appendix \ref{app:hcf}. We fed the features into a random forest classifier with $100$ decision trees. During training, no maximum depth was specified, and all nodes were expanded until all leaves were pure. For the car's data set, the resulting decision tree estimators had an average depth of $29.19$, a minimum depth of $25$, and a maximum depth of $38$. For the e-scooter's data set, the resulting decision tree estimators had an average depth of $37.87$, a minimum depth of $31$, and a maximum depth of $48$. {The run time of the model was evaluated on the same machine achieving a 40 milliseconds per model evaluation. This significant increase in computation time with respect to the CNN-based model is attributed to the large set of handcrafted features necessary to calculate for each input. Both approaches can be improved via model and feature optimization which opens a direction for future work. In addition, the raw IMU data processing as described in Section II.C. adds several operations of matrix multiplication which are negligible in comparison with the evaluation time of the trained models. Similarly, classical dead reckoning approaches use similar data processing procedure with the addition of double integration with respect to time which is also negligible when compared to the trained models.}

\begin{figure*}[!ht]
    \centering
    \begin{subfigure}[b]{0.9\textwidth}
    
      \includegraphics[width = \textwidth ,keepaspectratio]{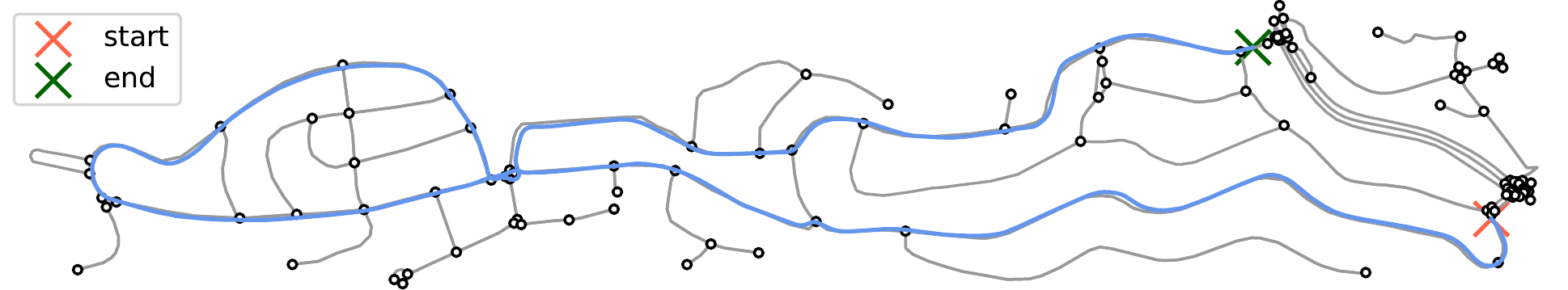}
        \caption{ }
        \label{figaa}

    \end{subfigure}%
    \\
       \begin{subfigure}[b]{0.9\textwidth}
      \includegraphics[width = \textwidth ,keepaspectratio]{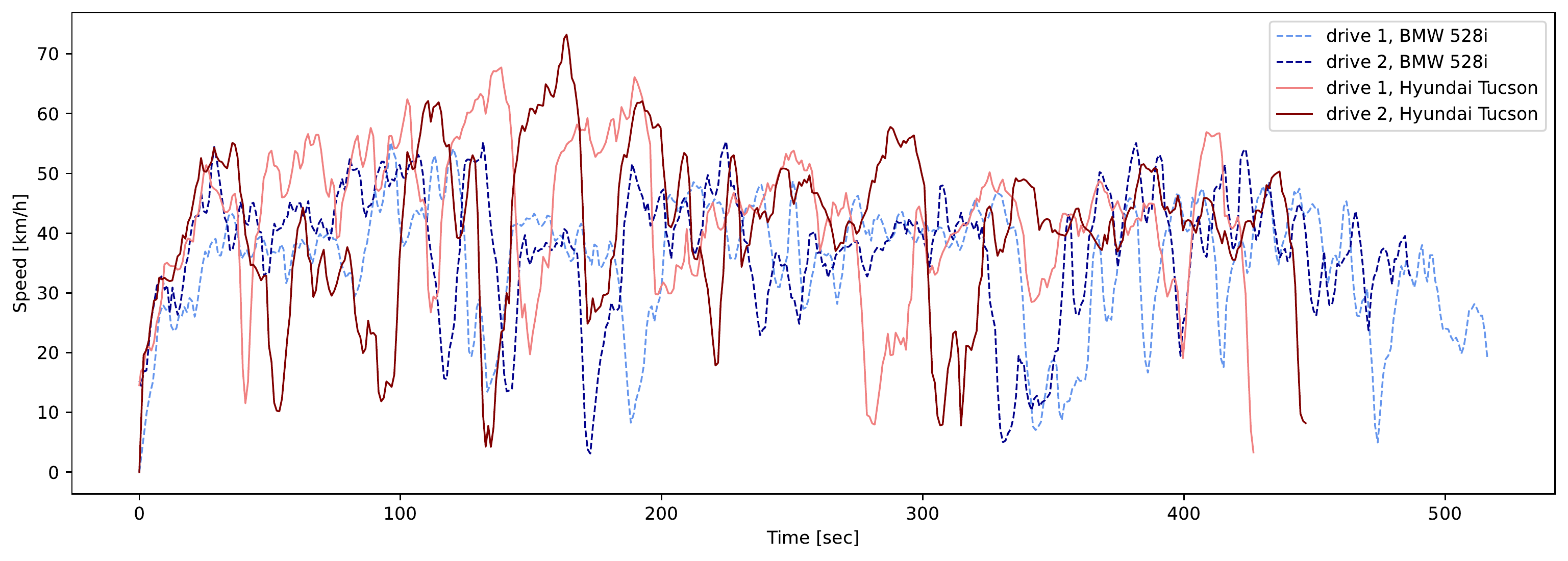} 
        \caption{}
        \label{figbb}
    \end{subfigure}%
    \hfill
   
    \caption{(a) The car route (blue) is shown on top of the road map (grey),  taken from OpenStreetMap \cite{bennett2010openstreetmap}., (b) Speed vs. time for four representative recordings with two different cars.} 
	\label{fig:pre_diagram}
\end{figure*}

\section{Experiments}
\label{sec:Experiments_data_collection}
We evaluated our positioning solution in two distinct scenarios: a car driving in a dense urban area and an e-scooter traveling on a route that includes both roads and sidewalks. As we detail in the sequel, the suggested data-driven methods significantly outperform traditional dead reckoning. In Subsection  \ref{sec:data_collection} we detail the data collected for both experiments, while the results of our approach are presented in Subsection \ref{sec:results}. 

\subsection{Data Collection}
\label{sec:data_collection}
We collected real-world data to construct two datasets suitable for testing and validating the proposed data-driven algorithm.

\subsubsection{Cars}
\label{sec:data_collection_cars}
We assembled a dataset consisting of 49 drives of a particular route, spanning a distance of 5919 [m], as seen in Figure \ref{fig:pre_diagram}. The data was recorded using the PhyPhox app \cite{staacks2018advanced} on a Galaxy A21s smartphone at a sampling rate of 200 Hz. {The mobile device was fixed between the driver and passenger seats at the same orientation in each drive.} In addition to the IMU data, GPS coordinates were also collected at a sampling rate of 1 Hz using the same app. The drives were taken in ten different cars, with five drives recorded from $9$ of the cars and $4$ drives from the 10th car. Two drives from each car were reserved for validation and testing. Therefore, a total of 29 drives were used for training, 10 for validation, and 10 for testing.
\subsubsection{E-scooters}
\label{sec:data_collection_scoots}
\begin{figure}[!ht]
    \centering
    \begin{subfigure}[b]{0.9\columnwidth}
      \includegraphics[width = \columnwidth ,keepaspectratio]{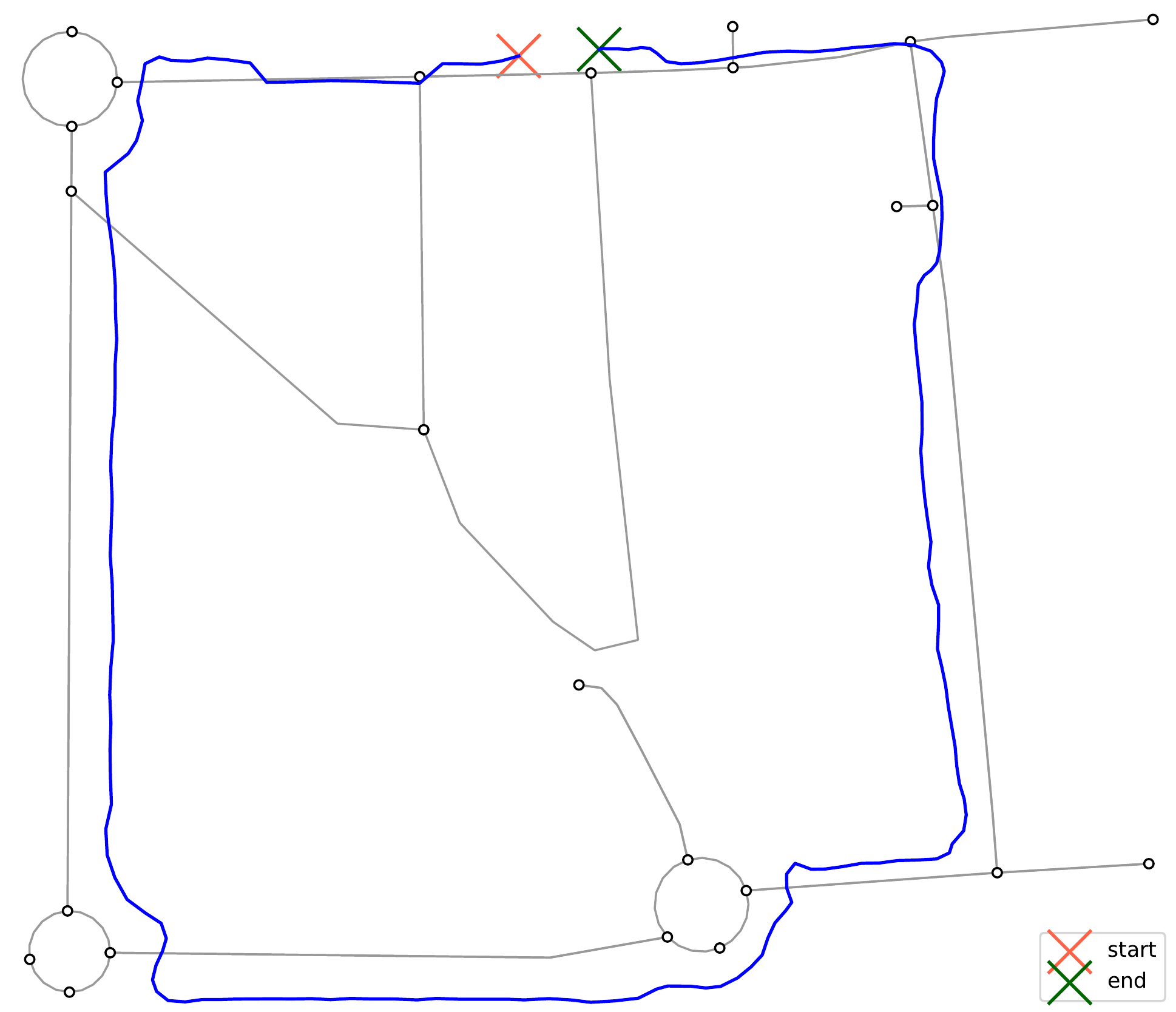}
        \caption{ }
        \label{figaa}
    \end{subfigure}%
    
    \begin{subfigure}[b]{1.0\columnwidth}
      \includegraphics[width = \columnwidth ,keepaspectratio]{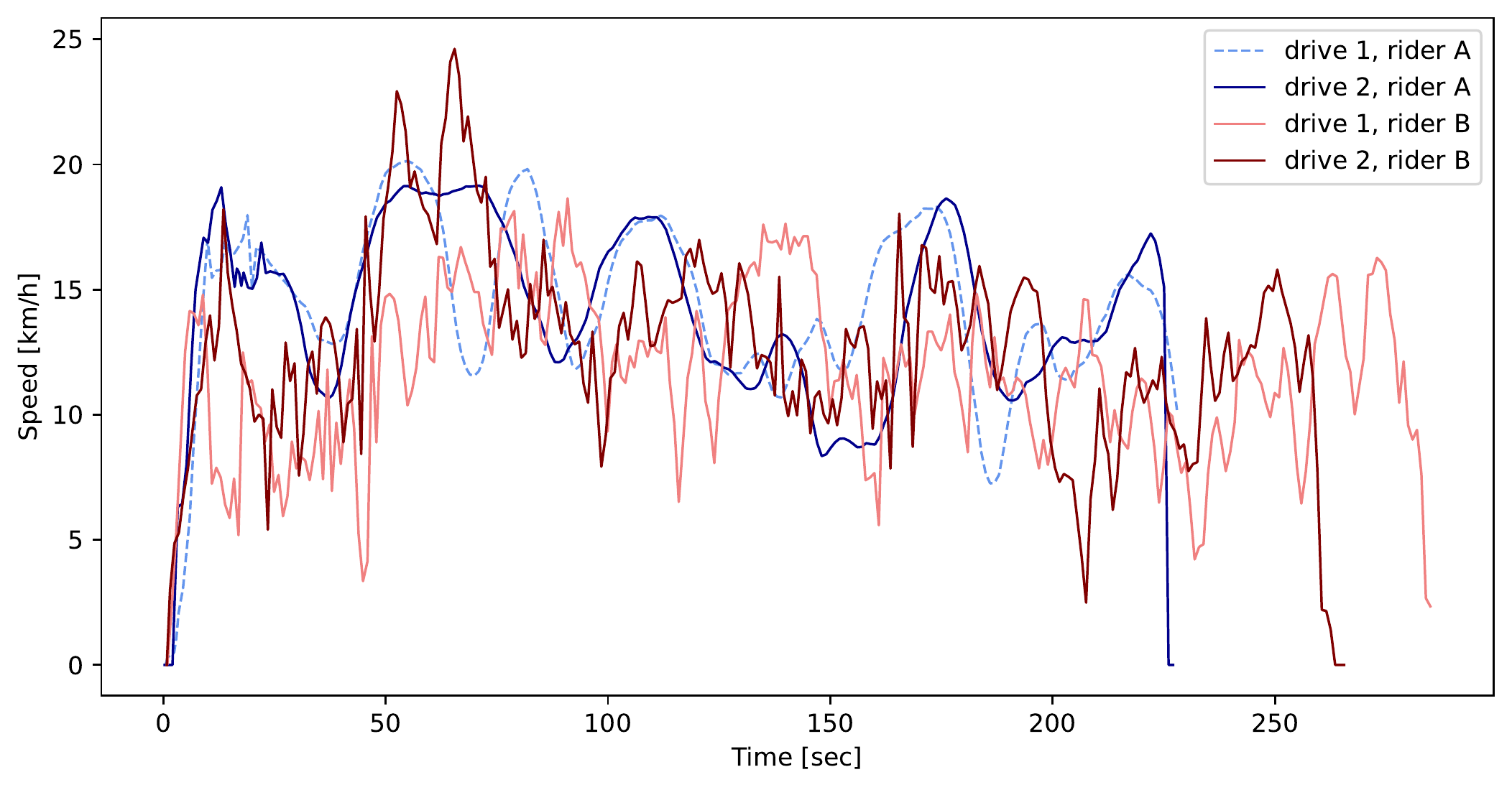} 
        \caption{}
        \label{figbb}
    \end{subfigure}%
    \hfill
   
    \caption{(a) The e-scooter route (blue) is shown on top of the road (grey), taken from OpenStreetMap. The route includes on and off-road segments, making for a diverse and challenging route, (b) Speed vs. time for four representative recordings with two different riders and phones. } 
	\label{fig:figmatam}
\end{figure}

We collected a data set of $63$ e-scooters drives covering a distance of $917 \ [m]$  on a predetermined route depicted in Figure \ref{fig:figmatam}. The drives were collected by two different drivers using different phones. The drivers were not told on which part of the route to drive on the sidewalk and on which on the road and each driver chose independently. The recordings contain IMU measurements at a frequency of $420$ Hz and GPS coordinates at a frequency of $1$ Hz recorded using the PhyPhox app on either a Galaxy S22 (for the first driver) or Oneplus nord2 (for the second driver) device. {The mobile devices were fixed using a phone holder firmly connected to the steering wheel.} We employed $47$ of the drives for training ($24$ from the first driver and $23$ from the second), $8$ for validation ($4$ from each driver), and $8$ for testing ($4$ from each driver).

\subsection{Experimental Results}
\label{sec:results}
This section presents the results of our experiments on cars and e-scooters, specifically evaluating the performance of both data-driven road segmentors described in Section \ref{sec:classifers}.

Our evaluation includes eight different measures, namely,
\begin{itemize}
    \item \textbf{acc raw} -- the raw accuracy of the classifier, i.e., using the road segmentor output.
    
    \item \textbf{acc} -- the accuracy of the classifier after applying the logic described in Algorithm \ref{ALG:inferposition}.
    
    \item \textbf{2-acc raw} -- the $2$ accuracy of the raw classifier, which is the percentage of segments at most $1$ segment removed from the ground truth.
    
    \item \textbf{2-acc} -- the $2$ accuracy after applying the logic in Algorithm \ref{ALG:inferposition}.
    
    \item \textbf{max dist raw} -- the maximal distance between the midpoint of the segment reported by the raw classifier and the ground truth.

    \item \textbf{max dist} -- the maximal distance between the position reported by Algorithm \ref{ALG:inferposition} and the ground truth.

    \item \textbf{mean dist raw} -- the average distance between the midpoint of the segment reported by the raw classifier and the ground truth.
    
    \item \textbf{mean dist} -- the average distance between the position reported by Algorithm \ref{ALG:inferposition} and the ground truth.
\end{itemize}
 The results of the evaluation are summarized in Table \ref{table:1} for cars and Table \ref{table:2} for e-scooters. All results were compared with classical dead reckoning where the IMU signals were rotated to the inertial frame and integrated with respect to time to yield location estimates.

\begin{table*}[ht]
\centering
\begin{tabular}{|c || c | c | c | c | c | c | c | c|} 
 \hline
 Classifier & acc raw & acc & 2-acc raw & 2-acc & max dist raw [m] & max dist [m] & mean dist raw [m] & mean dist [m] \\ 
 \hline\hline
 CNN & 0.79 & 0.87 & 0.89 & 1 & 2377  & 296  & 184 & 53 \\ 
 \hline
 Ensemble & 0.7 & 0.83 & 0.81 & 1 & 2377  & 211  & 198 & 55\\ 
 \hline
 Dead reckoning  & N.A. & N.A. & N.A. & N.A. & N.A. & 2484   & N.A. & 1247\\
 \hline
\end{tabular}
\caption{ Results of our data-driven approaches on the cars data-set. The performance metrics are explained in Section \ref{sec:results}.}
\label{table:1}
\end{table*}

\begin{table*}[!ht]
\centering
\begin{tabular}{|c || c | c | c | c | c | c | c | c|} 
 \hline
Classifier & acc raw & acc & 2-acc raw & 2-acc & max dist raw [m] & max dist [m] & mean dist raw [m] & mean dist [m] \\ 
 \hline\hline
 CNN & 0.58 & 0.72 & 0.70 & 0.96 & 297 & 149 & 74 & 32 \\ 
 \hline
 Ensemble & 0.48 & 0.63 & 0.63 &  0.9 & 316 & 309 &  97 & 56\\ 
 
 \hline
 Dead reckoning  & N.A. & N.A. & N.A. & N.A. & N.A. & 752  & N.A. & 316\\
 \hline
\end{tabular}
\caption{  Results on the e-scooters data-set.}
\label{table:2}
\end{table*}

\subsubsection{Learning car position}\label{sec:resultscars}
\begin{figure*}[!ht]
    \centering
    \begin{subfigure}[b]{0.9\columnwidth}
      \includegraphics[width = \columnwidth ,keepaspectratio]{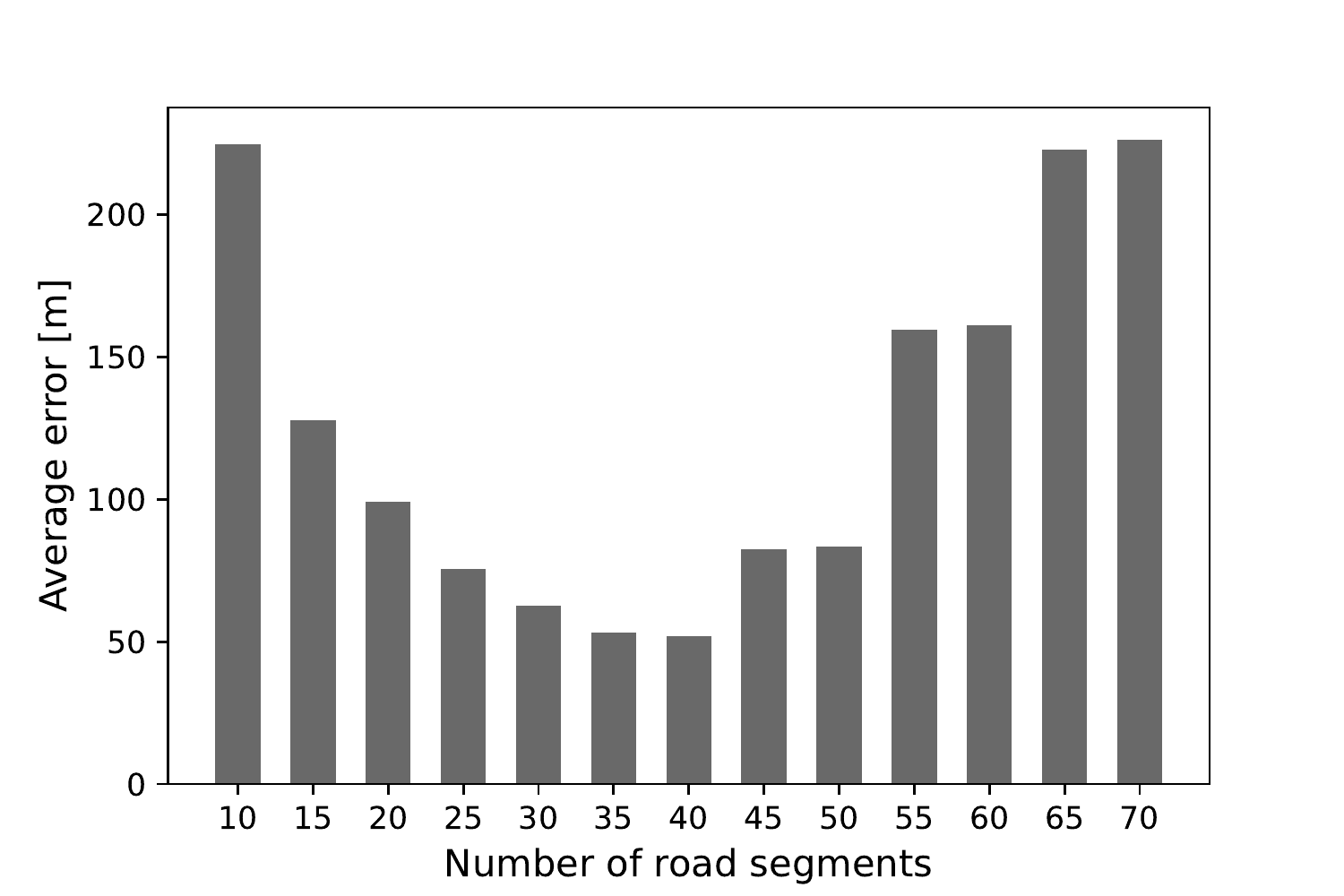}
        \caption{ }
        \label{figaa}
    \end{subfigure}%
       \begin{subfigure}[b]{0.9\columnwidth}
      \includegraphics[width = \columnwidth ,keepaspectratio]{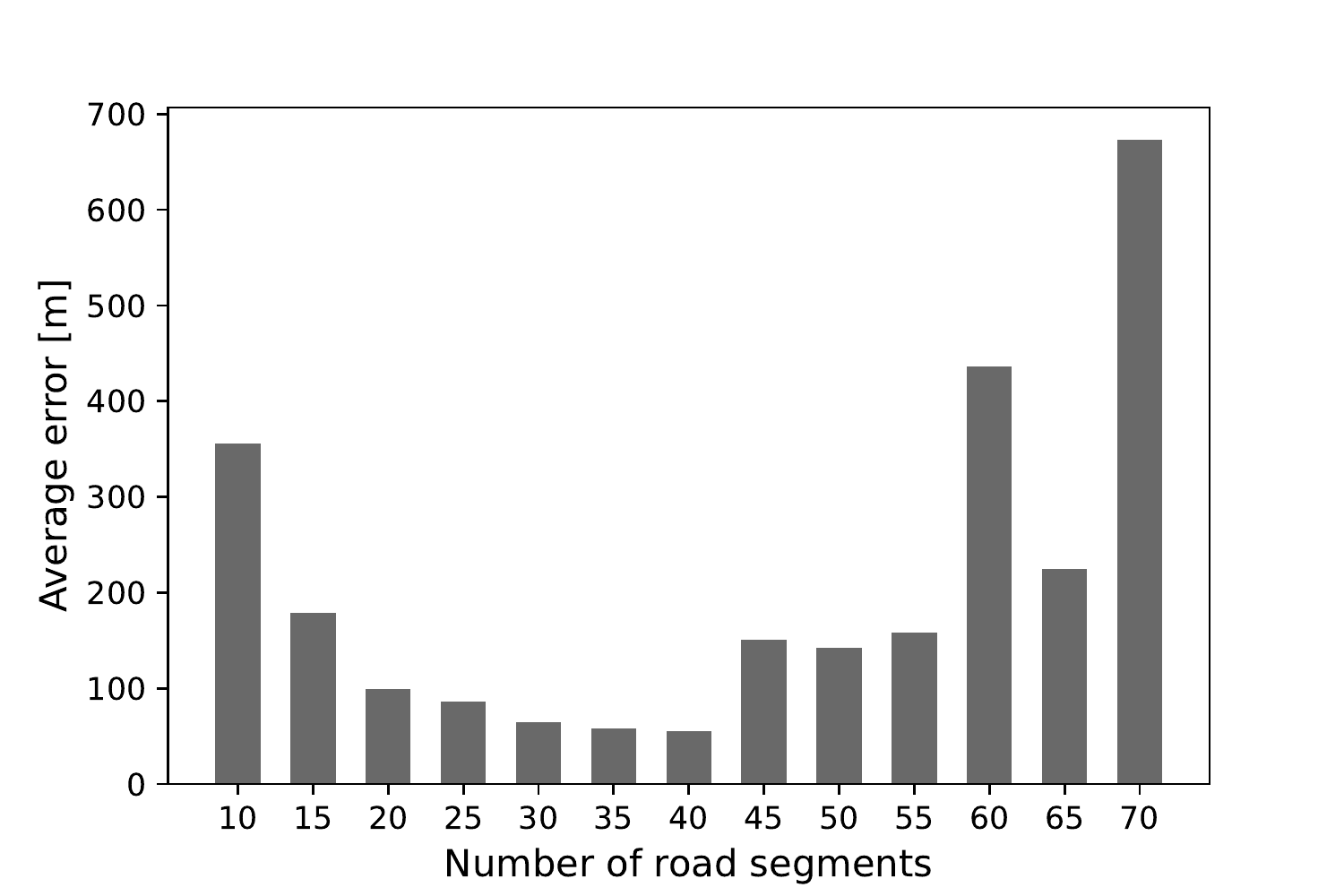} 
        \caption{}
        \label{figbb}
    \end{subfigure}%
    \hfill
   
    \caption{Average distance error [m] vs. the number of road segments, evaluated on the validation set for cars. The results are shown for both a CNN classifier (on the left) and a classifier based on hand-crafted features and a random forest (on the right).} 
	\label{fig:num_segs}
\end{figure*}

 To determine the optimal number of road segments for Algorithm \ref{ALG:inferposition}, we evaluated 13 different numbers of road segments, ranging from 10 to 70 in increments of 5, as described in Subsection \ref{sec:num_segments}. The optimal number of road segments was chosen based on the best performance on the validation set. Figure \ref{fig:num_segs} shows the resulting average error between the output of Algorithm \ref{ALG:inferposition} and the GPS  as a function of the number of segments for both the CNN and Ensemble road segmentors. We found that dividing the route into 40 segments yielded the lowest error on the validation set for both road segmentors.

\begin{figure*}[!ht]
    \centering
    \begin{subfigure}[b]{0.9\columnwidth}
      \includegraphics[width = \columnwidth ,keepaspectratio]{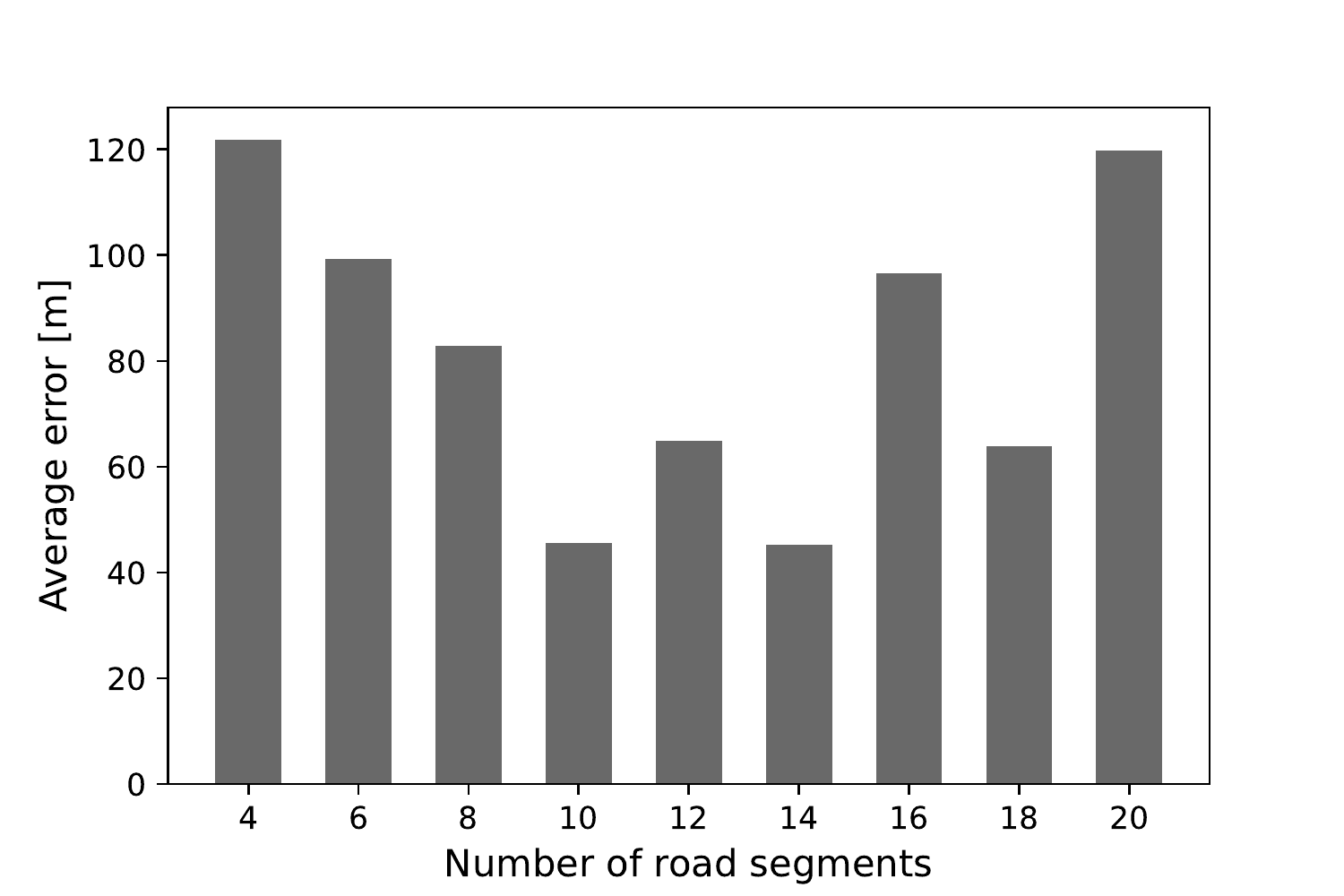}
        \caption{ }
        \label{figaa_matam}
    \end{subfigure}%
       \begin{subfigure}[b]{0.9\columnwidth}
      \includegraphics[width = \columnwidth ,keepaspectratio]{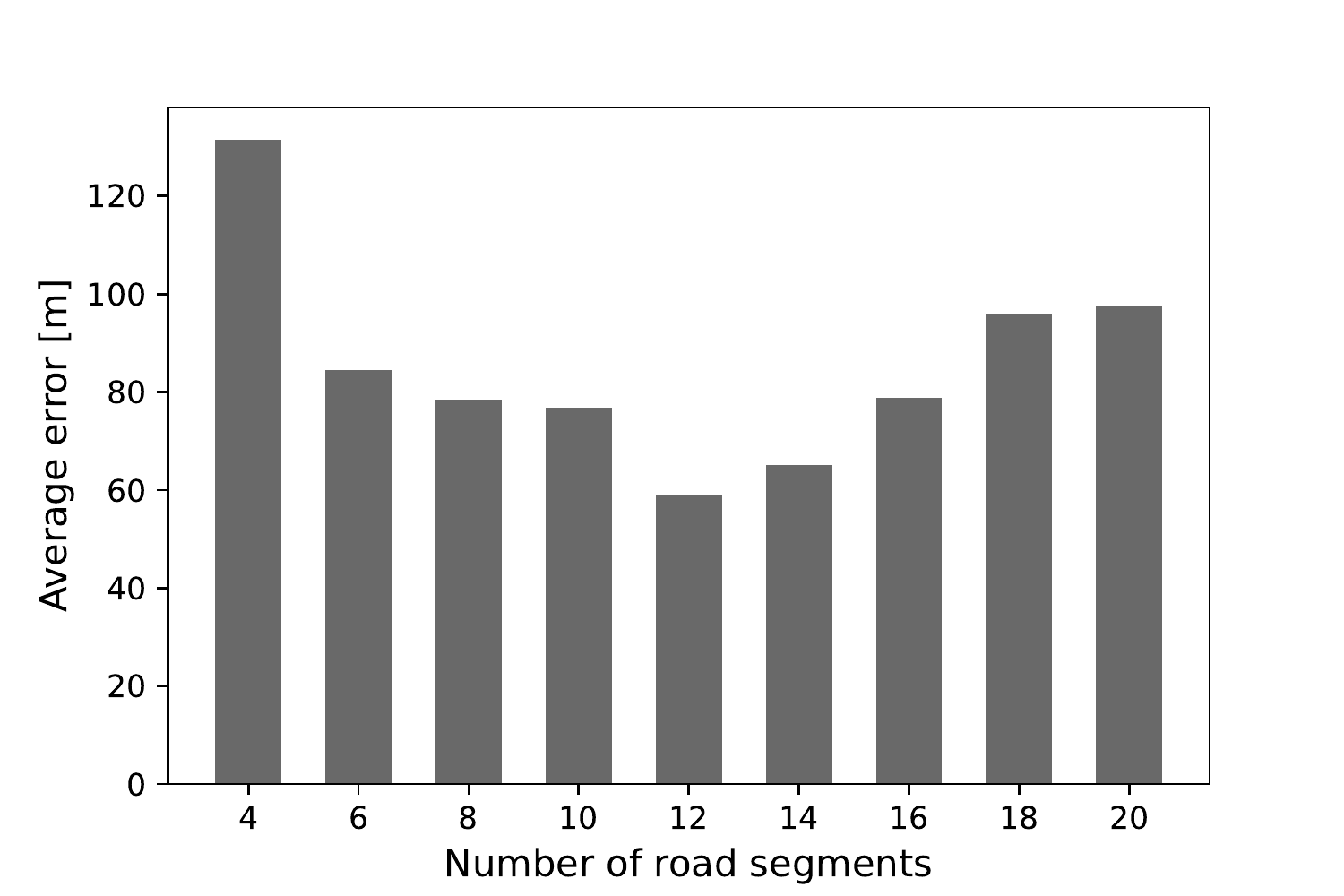} 
        \caption{}
        \label{figbb_matam}
    \end{subfigure}%
    \hfill
   
    \caption{Average distance error [m] vs. the number of road segments, evaluated on the validation set for e-scooters. The results are shown for both a CNN classifier (on the left) and a classifier based on hand-crafted features and a random forest (on the right).} 
	\label{fig:num_segs_matam}
\end{figure*}

\begin{figure*}[ht]
    \centering
    \begin{subfigure}[b]{0.99\columnwidth}
      \includegraphics[width = \columnwidth ,keepaspectratio]{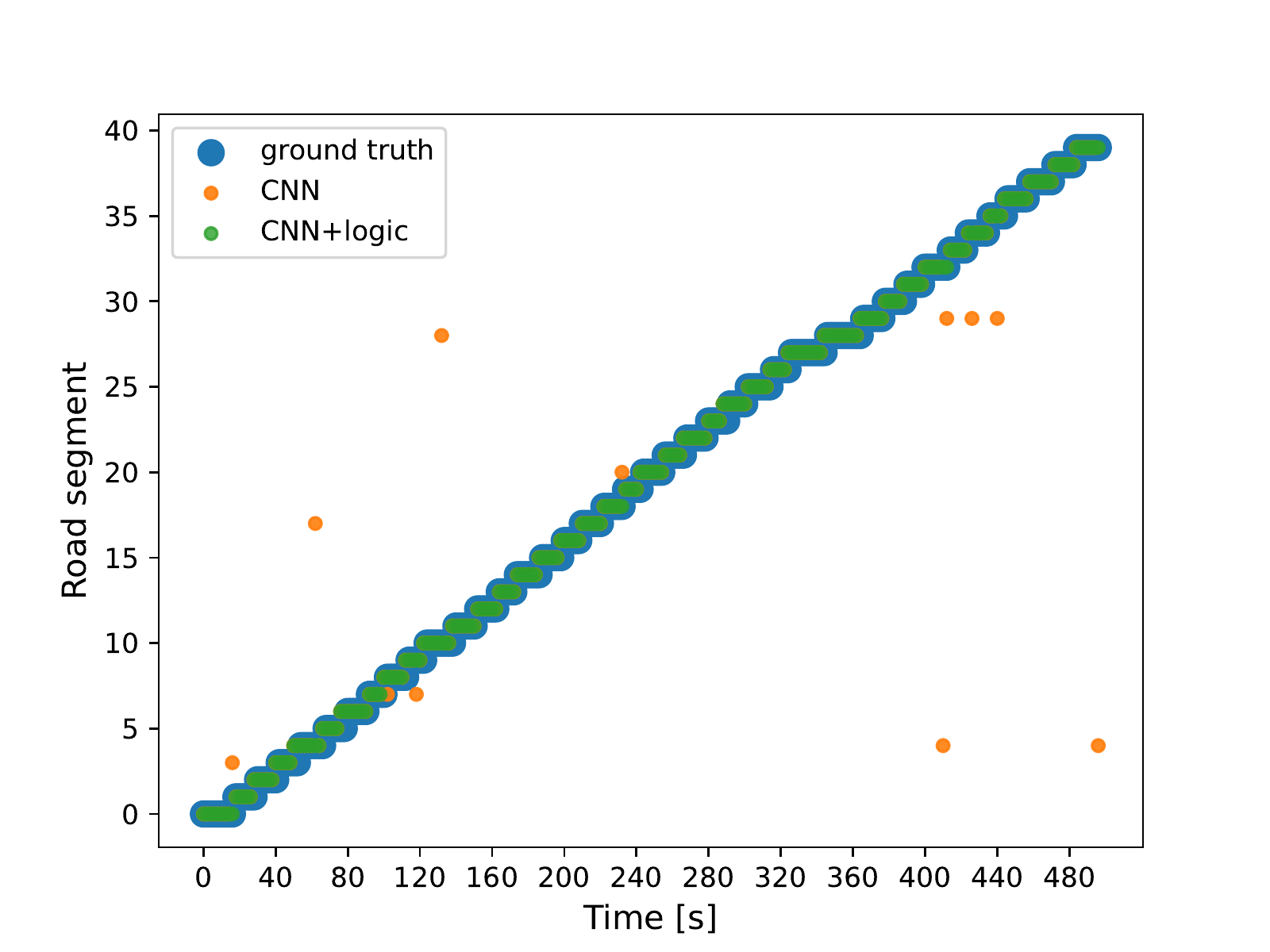}
        \caption{ }
        \label{figresltsa}
    \end{subfigure}%
       \begin{subfigure}[b]{0.99\columnwidth}
      \includegraphics[width = \columnwidth ,keepaspectratio]{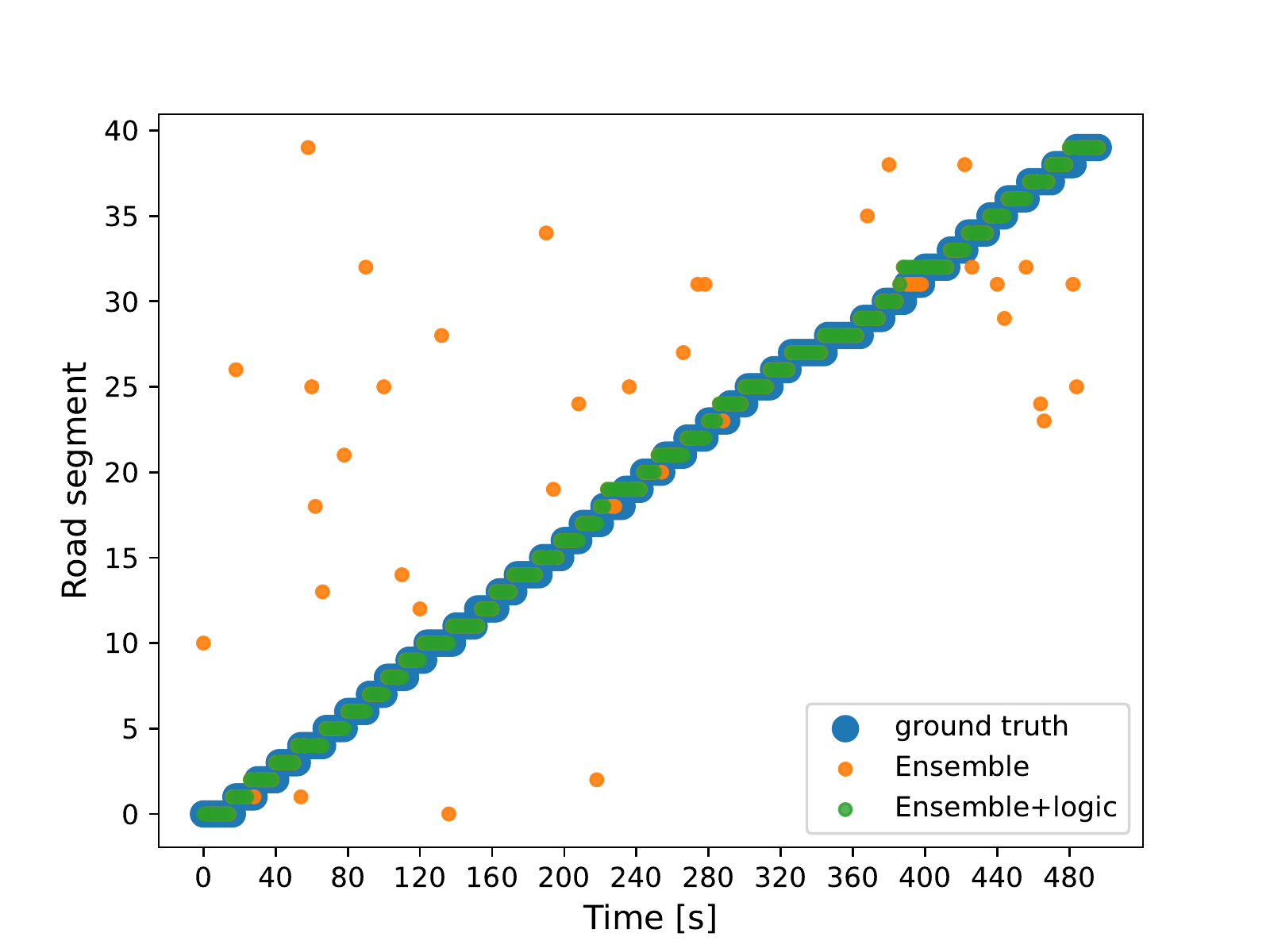}
        \caption{}
        \label{figresltsb}
    \end{subfigure}
    \\
        \begin{subfigure}[b]{0.99\columnwidth}
      \includegraphics[width = \columnwidth ,keepaspectratio]{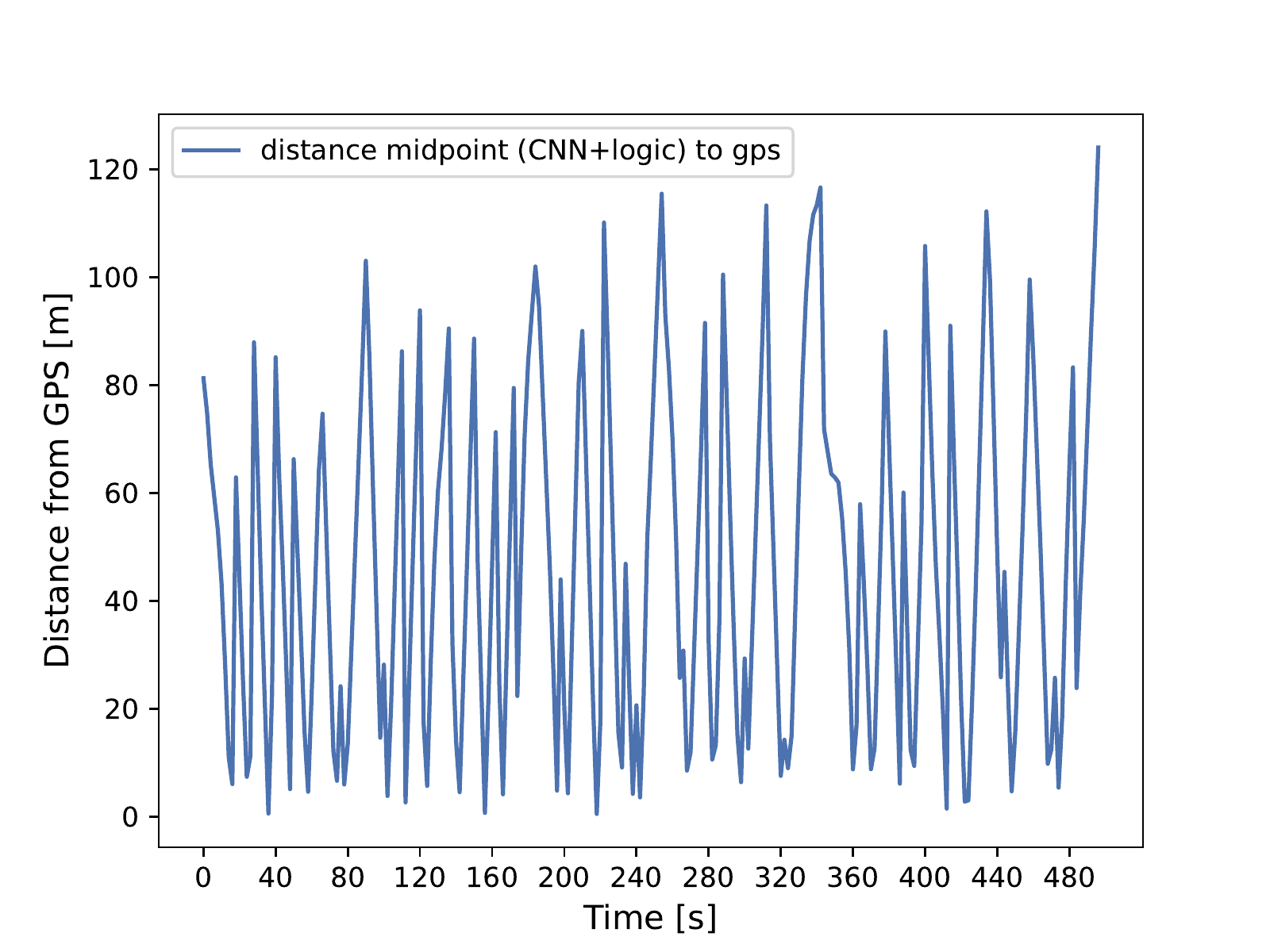}
        \caption{ }
        \label{figresltsc}
    \end{subfigure}%
       \begin{subfigure}[b]{0.99\columnwidth}
      \includegraphics[width = \columnwidth ,keepaspectratio]{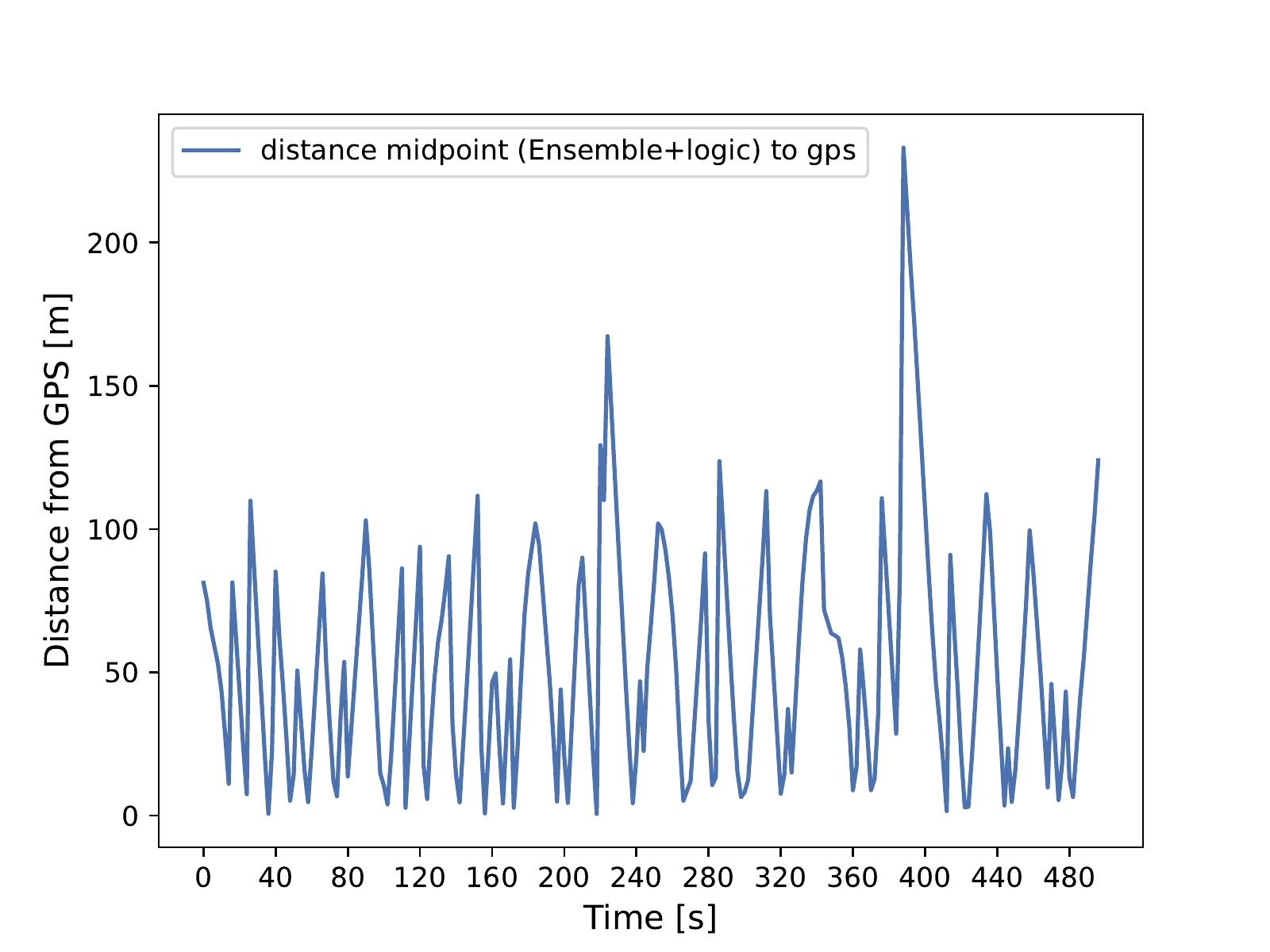}
        \caption{}
        \label{figresltsd}
    \end{subfigure}
    \caption{Results vs. time on a particular drive from the cars test set. The top row shows the classification results obtained by the DL-based road segmentor (CNN) and the road segmentor based on the ensemble of hand-crafted features and random forest (Ensemble). Both raw and corrected classification results are presented. The bottom row shows the error between the reported position and the GPS ground truth.} 
	\label{fig:results_cars}
\end{figure*}

\begin{figure*}[!ht]
    \centering
    \begin{subfigure}[b]{0.99\columnwidth}
      \includegraphics[width = \columnwidth ,keepaspectratio]{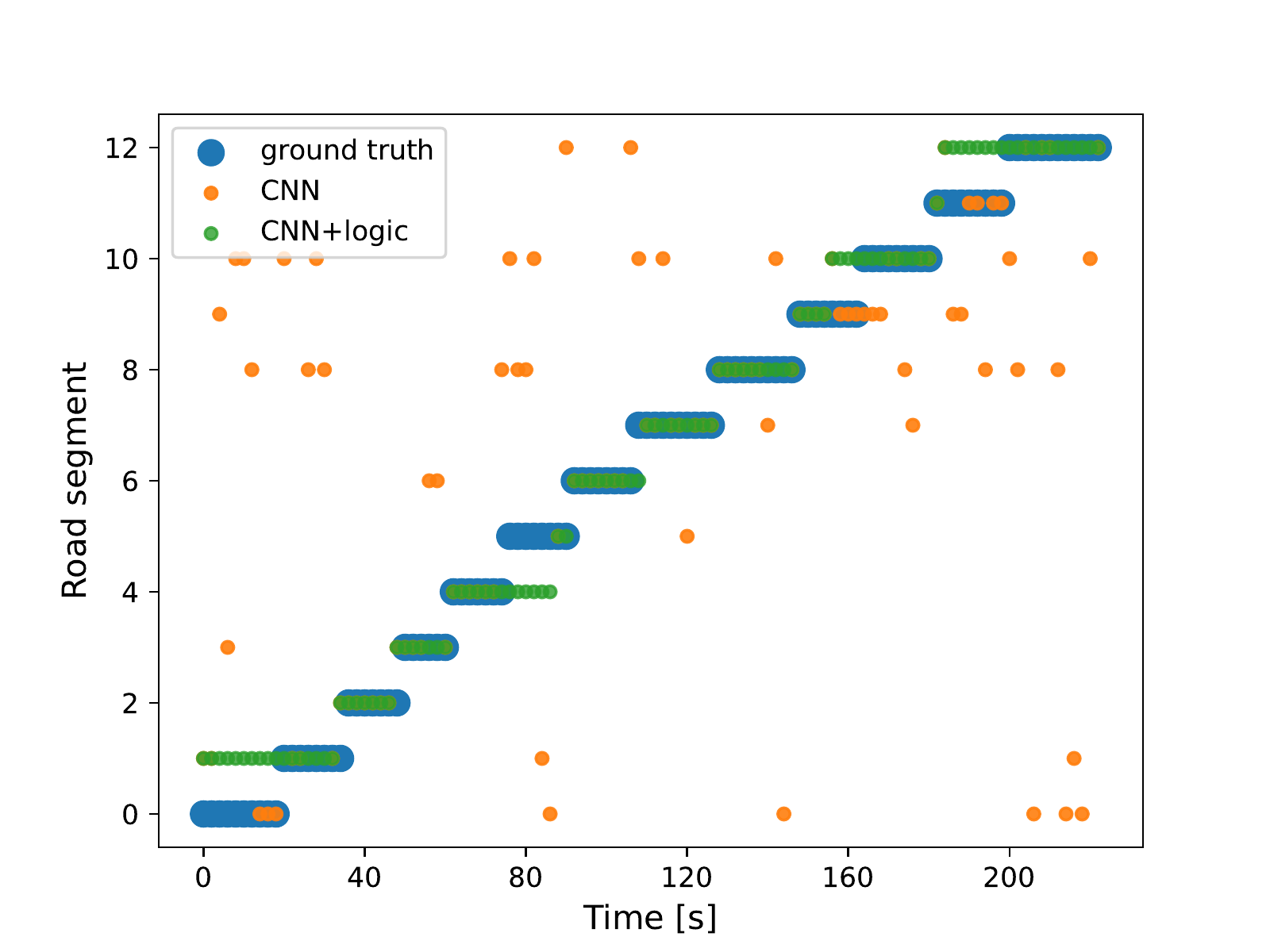}
        \caption{ }
        \label{figreslts1a}
    \end{subfigure}%
       \begin{subfigure}[b]{0.99\columnwidth}
      \includegraphics[width = \columnwidth ,keepaspectratio]{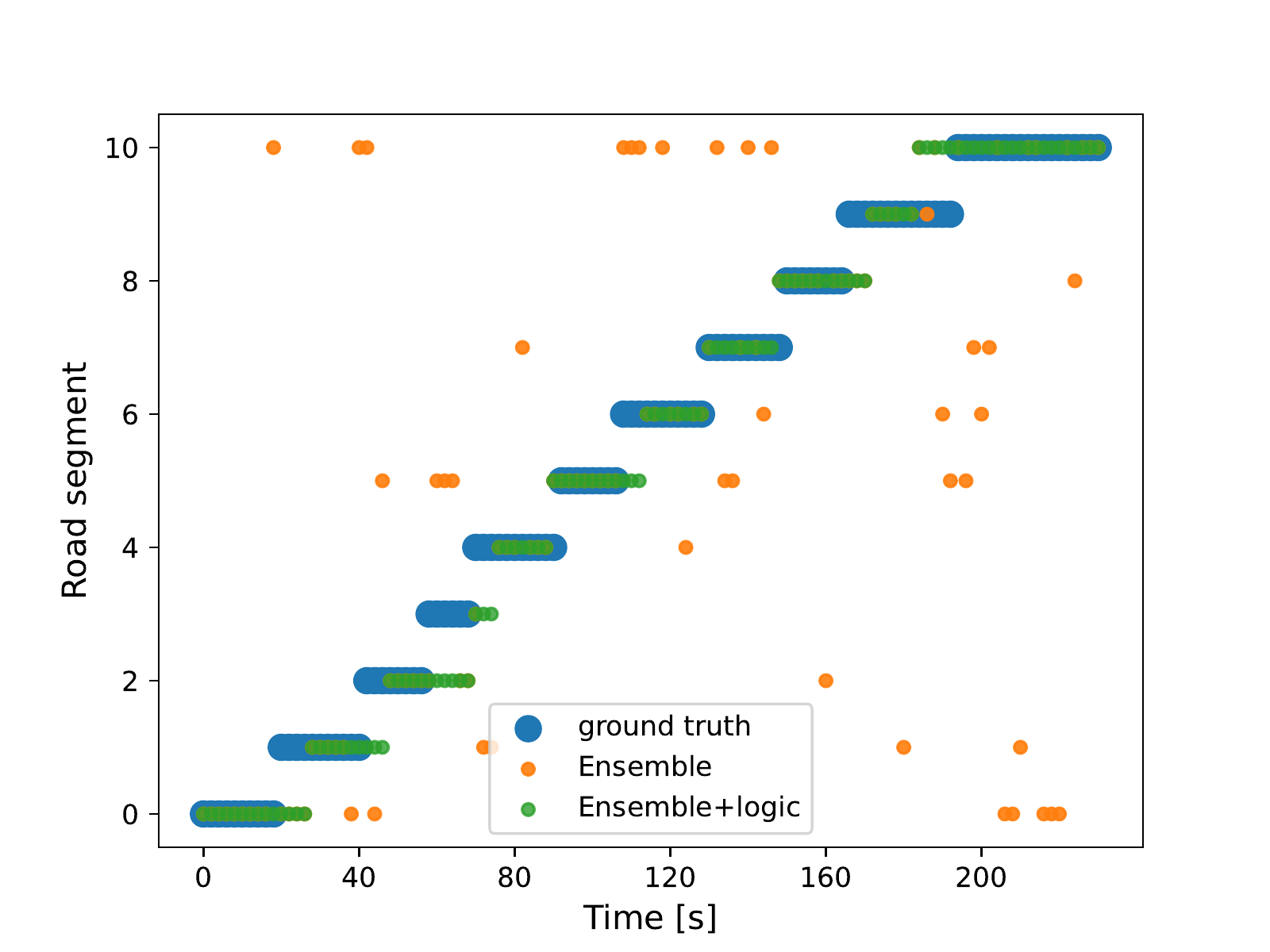}
        \caption{}
        \label{figreslts1b}
    \end{subfigure}
    \\
        \begin{subfigure}[b]{0.99\columnwidth}
      \includegraphics[width = \columnwidth ,keepaspectratio]{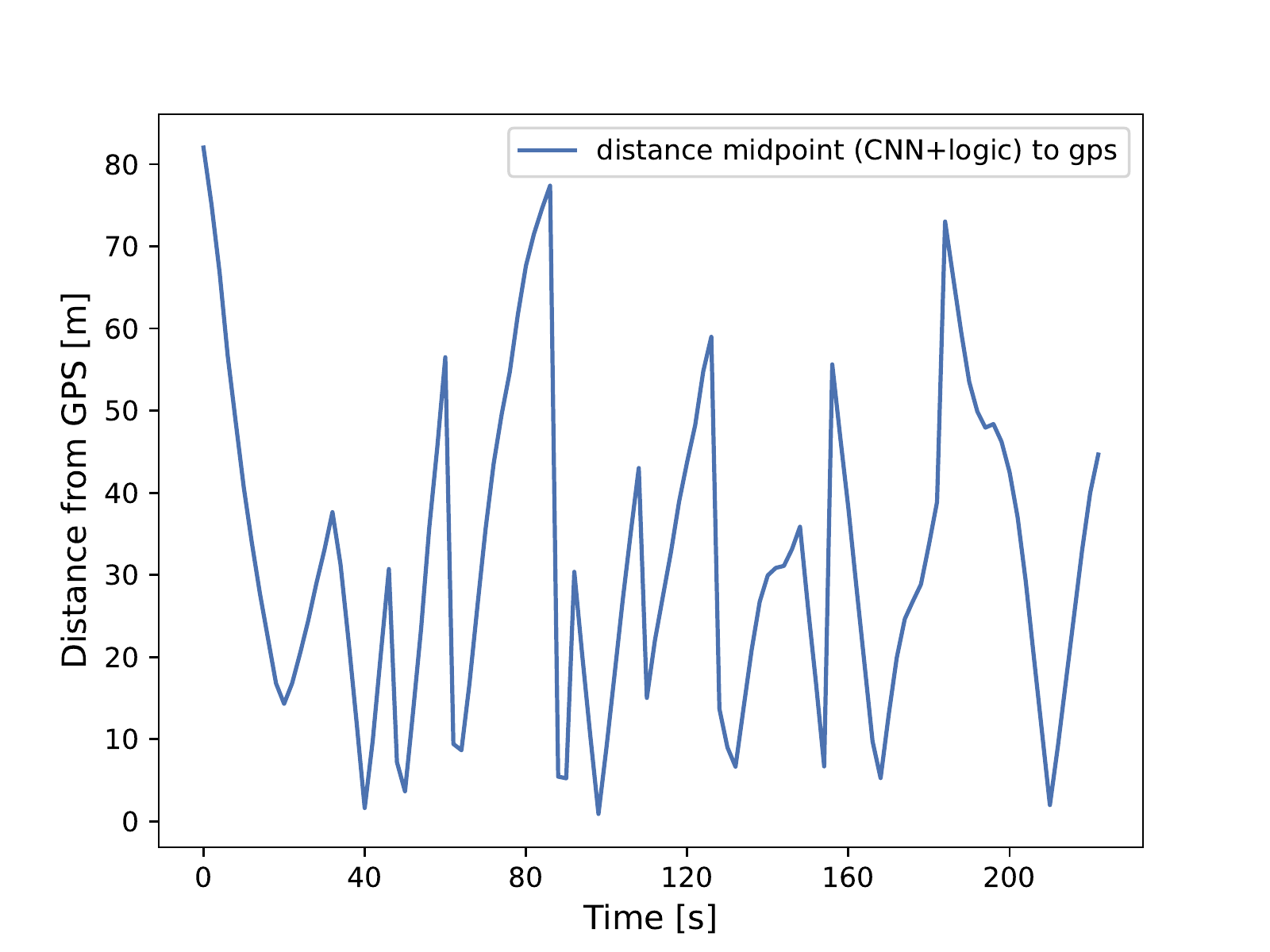}
        \caption{ }
        \label{figreslts1c}
    \end{subfigure}%
       \begin{subfigure}[b]{0.99\columnwidth}
      \includegraphics[width = \columnwidth ,keepaspectratio]{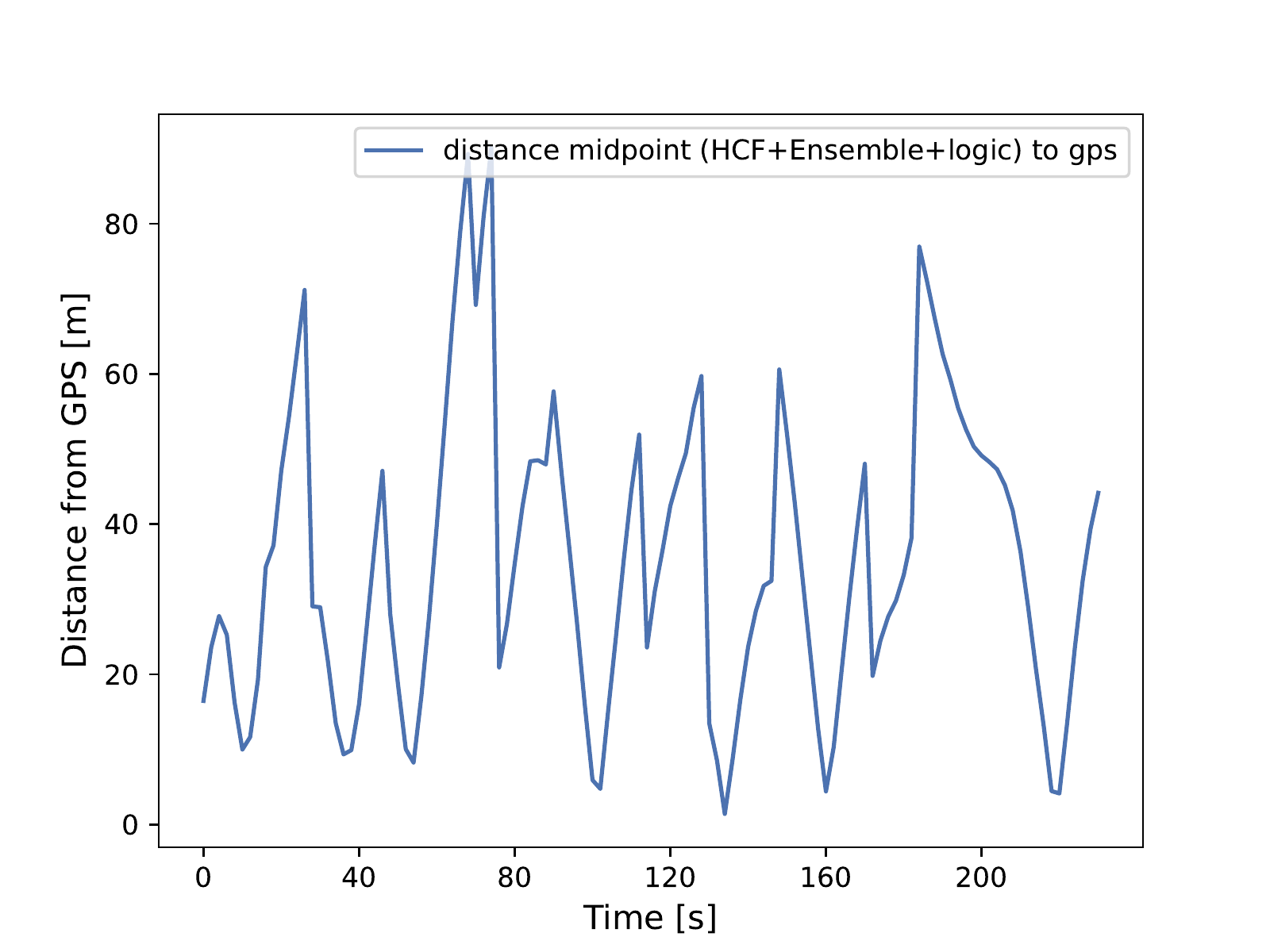}
        \caption{}
        \label{figreslts1d}
    \end{subfigure}
    \caption{Results vs. time on a particular drive from the e-scooter test set. The top row shows the classification results obtained by the DL-based road segmentor (CNN) and the road segmentor based on the ensemble of hand-crafted features and random forest (Ensemble). Both raw and corrected classification results are presented. The bottom row shows the error between the reported position and the GPS ground truth.} 
	\label{fig:results_scoots}
\end{figure*}

\subsubsection{Learning e-scooter position}\label{sec:resultsscoots}
Learning road segments from the road vibration generated by an e-scooter is more challenging than those generated by a car. This is due to the greater freedom an e-scooter driver has in choosing a route, resulting in a more varied road vibration signature. While a car completely fills a car lane, an e-scooter may drive on various paths on a sidewalk or road or a bicycle lane. Note the trajectory in our experiment in Figure \ref{fig:figmatam} where the e-scooter drives both off and on the road. In addition, an e-scooter will be more sensitive to turns, starts, and stops, resulting in noisier IMU measurements. As a result, the road vibration signature of an e-scooter is much more varied. Despite the challenges inherent in learning road segments from the road vibrations generated by an e-scooter, our proposed approach achieved an average error of approximately $30 \ [m]$ for a $917 \ [m]$ route. In contrast, a traditional dead reckoning approach resulted in an average error exceeding $300 \ [m]$. Comprehensive results of our positioning solution for the e-scooter dataset are provided in Table \ref{table:2}. 

We consider $9$ possibilities for the number of segments  between $4$ and $20$ with increments of $2$. For each, we partition the route into equal segments, train a road segmentor on the training set, and test the accuracy on the validation set using Algorithm \ref{ALG:inferposition}. In Figure \ref{fig:num_segs_matam}, the average error on the validation set is plotted as a function of the number of segments for both a deep classifier and a road segmentor based on random forest and HCFs. The optimal number of segments was found to be $14$ for the CNN road segmentor, and $12$ for the Ensemble road segmentor based on the lowest validation error.

\subsection{Performance Analysis}
To gain insight into the effectiveness of our approach, we can compare the maximum and average errors presented in Tables \ref{table:1} and \ref{table:2} to those achievable with a perfect road segmentor and a vehicle traveling at a constant velocity. If a route of length $L$ is partitioned into $N$ segments, a perfect road segmentor would have a maximum error of $\frac{L}{2N}$. Assuming constant speed, a perfect road segmentor would achieve an average error of $\frac{L}{4N}$. To see this, let $\bar{d}$ be the average distance between the vehicles position and the midpoints of the segments under the constant velocity assumption, $v$ be the constant velocity, and $T=\frac L{2Nv}$ be the time it takes to travel half a segment. We will then have,
\begin{equation}
    \bar{d} = \frac1T \intop_{t=0}^{T} \left[\frac{L}{2N}-vt \right]dt =\frac{L}{2N} - \frac12vT 
\end{equation}
plug in $vT=\frac{L}{2N}$ and we get $\bar{d} = \frac{L}{4N}$. Simply put, when traveling at constant velocity the average distance from the midpoint of the segment is equal to one-quarter of the segment's length.

In the car experiments, the route length was $5919 \ [m]$, divided into $40$ segments. This suggests a maximal error of $73 \ [m]$ and an average error of $36 \ [m]$ with a perfect road segmentor. However, our approach resulted in a maximal error of $184 \ [m]$ and $198 \ [m]$ and an average error of $53 \ [m]$ and $55 \ [m]$ for the CNN and Ensemble road segmentors, respectively, which are close to the ideal values. In comparison to the classical dead reckoning method, our results demonstrate that the maximal error from our road segmentors is over ten times smaller, and the average error is over twenty times smaller. Full details can be found in Table \ref{table:1}.

In the e-scooter experiment, the route spanned $917 \ [m] $  and was divided into 14 segments for the CNN road segmentor and 12 segments for the Ensemble road segmentor. A perfect road segmentor would attain a maximal error of $32 \ [m]$ and $38 \ [m]$  for the CNN and Ensemble, respectively, with an expected average error of about $16 \ [m]$  and $19 \ [m] $ for the respective segmentors. Our experiment produced a maximal error of $149 \ [m]$ and $309 \ [m] $  for the CNN and Ensemble segmentors, respectively, with an average error of $32 \ [m] $   and $56 \ [m] $ for the CNN and Ensemble, respectively. Our results are roughly the same order of magnitude as those of the optimal road segmentor, despite the difficulty of the e-scooter localization task. Moreover, our data-driven approaches improve the mean distance error 5 times over classical dead reckoning.
 
The significance of the wrapping logic function $L$ in Eq. \ref{eq:logic} should be noted. As shown in Tables \ref{table:1} and \ref{table:2}, the accuracy of both road segmentation and position estimation is significantly improved with the inclusion of the logic function $L$. This improvement is further evident in Figures \ref{fig:results_cars} and \ref{fig:results_scoots}, where the unprocessed road segmentation (depicted in orange) appears noisy and inaccurate. However, the accuracy is significantly enhanced upon application of the logic function $L$, which eliminates severe outliers. For instance, in the car dataset, the maximum distance without logic is more than ten times larger.

\section{Conclusion} \label{sec:conclusions}

\subsection{Towards GPS-independent positioning solutions}

This work proposes a novel approach for vehicle positioning that does not rely on the GNSS. GNSS approaches are vulnerable to interference or failure in certain environments, rendering them unreliable in many situations. To address this issue, the proposed approach learns the road signature of a given environment and uses it for positioning via vehicle vibration, using only the IMU sensor. The approach divides a route into segments, each with a distinct signature that the IMU can detect by identifying subtle changes in the road surface. The study introduces two different methods for learning the road segment from IMU measurements, one based on convolutional neural networks and the other on ensemble random forest applied to handcrafted features. An algorithm for estimating the vehicle's road segment and deducing position in real time was presented.

The proposed approach was evaluated on a $6 \ [km]$ route in a dense urban area for a car and a $1 \ [km]$ route that combined road and pavement surfaces for an e-scooter. The mean error between the proposed method's position and the ground truth was approximately $50 \ [m]$  for the car and $30 \ [m] $ for the e-scooter, representing significant improvements over the classical dead reckoning approach. The findings demonstrate the potential for effective and efficient vehicle positioning techniques that can be used in situations where traditional GNSS approaches are unreliable.

\subsection{Future Work}
{This work opens the door to several directions for future research: (1) improving the machine learning methodology for more accurate positioning from a given window of IMU measurements, (2) improving the navigation scheme to use the vibration-based position updates with other sources (odometry, cameras, LiDAR, etc.), and (3) improving the system as a whole.

The first and most clear direction is to improve the segment classification accuracy. For example, investigating the quality of different road segmentors based on different deep learning architectures or when the road segments are not of uniform length but divided by some other criteria. It is also interesting to consider an unsupervised or semi-supervised setting where the road segment is inferred solely from the IMU reading without GPS labeling. Such a setting is relevant to indoor navigation where GPS is not available and obtaining accurate labels is difficult. The second important research direction includes the fusion strategy of the present approach with other sensors such as cameras or LiDAR, combining the practically unlimited availability of the IMU with the benefits of other sensors. 

The last bullet refers to the study of the positioning system as a whole including robustness to changes. A system may be developed and studied that carries out data crowdsourcing for continuous model creation of unknown roads, refinement of known roads, and adaptation to changes in existing roads. However, it is important to note that in reality roads are dynamic - they get damaged and fixed over time. This may introduce unexpected behaviour in vibration-based positioning approaches and requires dedicated robustness studies.}

\section*{Appendix -- handcrafted features for ensemble random forest}
As discussed in Section \ref{sec:classifers}, one of the road segmentors considered in this work was an ensemble random forest with hand-crafted features. Here the particular features manually computed from the processed IMU signals are detailed.
\label{app:hcf}
 From the initial $6$ channels, $\omega_x, \omega_y, \omega_y, a_x, a_y, a_z$
of processed IMU data we generate additional $42$ channels. The numerical derivatives, $\omega_x^d, \omega_y^d, \omega_y^d, a_x^d, a_y^d, a_z^d$
the numerical  integrals, $\omega_x^i, \omega_y^i, \omega_y^i, a_x^i, a_y^i, a_z^i$
the quotients, $\frac{\omega_x}{\omega_y}, \frac{\omega_y}{\omega_x}, \frac{\omega_z}{\omega_x}, \frac{\omega_x}{\omega_z}, \frac{\omega_y}{\omega_z}, \frac{\omega_z}{\omega_y}, \frac{a_x}{a_y}, \frac{a_y}{a_x}, \frac{a_x}{a_z}, \frac{a_z}{a_x}, \frac{a_y}{a_z}, \frac{a_z}{a_y}$
the quotients of the numerical integrals,
$
\frac{\omega_x^i}{\omega_y^i}, \frac{\omega_y^i}{\omega_x^i}, \frac{\omega_z^i}{\omega_x^i}, \frac{\omega_x^i}{\omega_z^i}, \frac{\omega_y^i}{\omega_z^i}, \frac{\omega_z^i}{\omega_y^i}, \frac{a_x^i}{a_y^i}, \frac{a_y^i}{a_x^i}, \frac{a_x^i}{a_z^i}, \frac{a_z^i}{a_x^i}, \frac{a_y^i}{a_z^i}, \frac{a_z^i}{a_y^i}
$
and the amplitude of the Fourier coefficients 
$\mathcal{F}\{\omega_x\}, \mathcal{F}\{ \omega_y\}, \mathcal{F}\{\omega_y\}, \mathcal{F}\{a_x\}, \mathcal{F}\{a_y\}, \mathcal{F}\{a_z\}.
$
We extracted various features from the 48 channels of our IMU measurements, except for the Fourier coefficients. The following $12$ features were computed: mean, standard deviation, minimum, maximum, median, median absolute deviation, second moment, skewness, kurtosis, interquartile range, spectral entropy, and largest value divided by smallest in absolute value. For the $24$ quotient channels, we also computed the mean absolute deviation.
For the $6$ channels of Fourier coefficients, we computed all of the features above, with the exception of spectral entropy, resulting in additional $66$ channels. 
We also computed the signal magnitudes area for the following triplets $\{\omega_x, \omega_y, \omega_z\}$ $\{a_x, a_y, a_z\}$,  $\{a_x^d, a_y^d, a_z^d\}$, $ \{\omega_x^d, \omega_y^d, \omega_z^d\} $, $\{a_x^i, a_y^i, a_z^i\}$, $\{\omega_x^i, \omega_y^i, \omega_z^i\}$, $\{\mathcal{F}\{\omega_x\}, \mathcal{F}\{ \omega_y\}, \mathcal{F}\{\omega_y\}\}$ and 
$\{\mathcal{F}\{a_x\}, \mathcal{F}\{a_y\}, \mathcal{F}\{a_z\}\}$ for additional $8$ channels.
Additionally, we computed $24$ correlation coefficients, between the following couplets $\{\omega_x, \omega_y\}, \{\omega_x, \omega_z\}$, $\{\omega_x, \omega_z\}$, $\{a_x, a_y\}$, $\{a_x, a_z\}$, $\{a_y, a_z\}$, $\{\omega_x^d, \omega_y^d\}, \{\omega_x^d, \omega_z^d\}$, $\omega_x^d, \omega_z^d$, $\{a_x^d, a_y^d\}$, $\{a_x^d, a_z^d\}$, $\{a_y^d, a_z^d\}$,
$\{\omega_x^i, \omega_y^i\}, \{\omega_x^i, \omega_z^i\}$, $\{\omega_x^i, \omega_z^i\}$, $\{a_x^i, a_y^i\}$, $\{a_x^i, a_z^i\}$, $\{a_y^i, a_z^i\}$, 
$\{\mathcal{F}\{a_x\}, \mathcal{F}\{a_y\}\}$, $\{\mathcal{F}\{a_y\}, \mathcal{F}\{a_z\}\}$, $\{\mathcal{F}\{a_x\}, \mathcal{F}\{a_z\}\}$, 
$\{\mathcal{F}\{\omega_x\}, \mathcal{F}\{ \omega_y\}\}$, $\{\mathcal{F}\{ \omega_y\}, \mathcal{F}\{\omega_y\}\}$, and  $\{\mathcal{F}\{\omega_x\} \mathcal{F}, \{\omega_y\}\}$.
Lastly, we employed an order 2 vector autoregressive model to fit the IMU measurements and appended the resulting coefficients as additional features.

\bibliographystyle{IEEEtran}
\bibliography{IEEEfull}

\begin{thebibliography}{10}
\providecommand{\url}[1]{#1}
\csname url@samestyle\endcsname
\providecommand{\newblock}{\relax}
\providecommand{\bibinfo}[2]{#2}
\providecommand{\BIBentrySTDinterwordspacing}{\spaceskip=0pt\relax}
\providecommand{\BIBentryALTinterwordstretchfactor}{4}
\providecommand{\BIBentryALTinterwordspacing}{\spaceskip=\fontdimen2\font plus
\BIBentryALTinterwordstretchfactor\fontdimen3\font minus \fontdimen4\font\relax}
\providecommand{\BIBforeignlanguage}[2]{{%
\expandafter\ifx\csname l@#1\endcsname\relax
\typeout{** WARNING: IEEEtran.bst: No hyphenation pattern has been}%
\typeout{** loaded for the language `#1'. Using the pattern for}%
\typeout{** the default language instead.}%
\else
\language=\csname l@#1\endcsname
\fi
#2}}
\providecommand{\BIBdecl}{\relax}
\BIBdecl

\bibitem{langley2017introduction}
R.~B. Langley, P.~J. Teunissen, and O.~Montenbruck, ``Introduction to gnss,'' \emph{Springer handbook of global navigation satellite systems}, pp. 3--23, 2017.

\bibitem{zangenehnejad2021gnss}
F.~Zangenehnejad and Y.~Gao, ``Gnss smartphones positioning: Advances, challenges, opportunities, and future perspectives,'' \emph{Satellite navigation}, vol.~2, pp. 1--23, 2021.

\bibitem{farrell2008aided}
J.~Farrell, \emph{Aided navigation: GPS with high rate sensors}.\hskip 1em plus 0.5em minus 0.4em\relax McGraw-Hill, Inc., 2008.

\bibitem{wahlstrom2016imu}
J.~Wahlstr{\"o}m, I.~Skog, P.~H{\"a}ndel, and A.~Nehorai, ``Imu-based smartphone-to-vehicle positioning,'' \emph{IEEE Transactions on Intelligent Vehicles}, vol.~1, no.~2, pp. 139--147, 2016.

\bibitem{el2021indoor}
N.~El-Sheimy and Y.~Li, ``Indoor navigation: State of the art and future trends,'' \emph{Satellite Navigation}, vol.~2, no.~1, pp. 1--23, 2021.

\bibitem{francis2020long}
A.~Francis, A.~Faust, H.-T.~L. Chiang, J.~Hsu, J.~C. Kew, M.~Fiser, and T.-W.~E. Lee, ``Long-range indoor navigation with prm-rl,'' \emph{IEEE Transactions on Robotics}, vol.~36, no.~4, pp. 1115--1134, 2020.

\bibitem{al2012indoor}
O.~Al~Hammadi, A.~Al~Hebsi, M.~J. Zemerly, and J.~W. Ng, ``Indoor localization and guidance using portable smartphones,'' in \emph{2012 IEEE/WIC/ACM International Conferences on Web Intelligence and Intelligent Agent Technology}, vol.~3.\hskip 1em plus 0.5em minus 0.4em\relax IEEE, 2012, pp. 337--341.

\bibitem{zhao2023data}
C.~Zhao, A.~Song, Y.~Zhu, S.~Jiang, F.~Liao, and Y.~Du, ``Data-driven indoor positioning correction for infrastructure-enabled autonomous driving systems: A lifelong framework,'' \emph{IEEE Transactions on Intelligent Transportation Systems}, 2023.

\bibitem{zhu2018gnss}
N.~Zhu, J.~Marais, D.~B{\'e}taille, and M.~Berbineau, ``Gnss position integrity in urban environments: A review of literature,'' \emph{IEEE Transactions on Intelligent Transportation Systems}, vol.~19, no.~9, pp. 2762--2778, 2018.

\bibitem{jing2022integrity}
H.~Jing, Y.~Gao, S.~Shahbeigi, and M.~Dianati, ``Integrity monitoring of gnss/ins based positioning systems for autonomous vehicles: State-of-the-art and open challenges,'' \emph{IEEE Transactions on Intelligent Transportation Systems}, 2022.

\bibitem{rizos2013locata}
C.~Rizos, ``Locata: A positioning system for indoor and outdoor applications where gnss does not work,'' in \emph{Proceedings of the 18th Association of Public Authority Surveyors Conference}, 2013, pp. 73--83.

\bibitem{yeong2021sensor}
D.~J. Yeong, G.~Velasco-Hernandez, J.~Barry, and J.~Walsh, ``Sensor and sensor fusion technology in autonomous vehicles: A review,'' \emph{Sensors}, vol.~21, no.~6, p. 2140, 2021.

\bibitem{meng2017robust}
X.~Meng, H.~Wang, and B.~Liu, ``A robust vehicle localization approach based on gnss/imu/dmi/lidar sensor fusion for autonomous vehicles,'' \emph{Sensors}, vol.~17, no.~9, p. 2140, 2017.

\bibitem{ziebinski2016survey}
A.~Ziebinski, R.~Cupek, H.~Erdogan, and S.~Waechter, ``A survey of adas technologies for the future perspective of sensor fusion,'' in \emph{Computational Collective Intelligence: 8th International Conference, ICCCI 2016, Halkidiki, Greece, September 28-30, 2016. Proceedings, Part II 8}.\hskip 1em plus 0.5em minus 0.4em\relax Springer, 2016, pp. 135--146.

\bibitem{campbell2018sensor}
S.~Campbell, N.~O'Mahony, L.~Krpalcova, D.~Riordan, J.~Walsh, A.~Murphy, and C.~Ryan, ``Sensor technology in autonomous vehicles: A review,'' in \emph{2018 29th Irish Signals and Systems Conference (ISSC)}.\hskip 1em plus 0.5em minus 0.4em\relax IEEE, 2018, pp. 1--4.

\bibitem{ibisch2013towards}
A.~Ibisch, S.~St{\"u}mper, H.~Altinger, M.~Neuhausen, M.~Tschentscher, M.~Schlipsing, J.~Salinen, and A.~Knoll, ``Towards autonomous driving in a parking garage: Vehicle localization and tracking using environment-embedded lidar sensors,'' in \emph{2013 IEEE intelligent vehicles symposium (IV)}.\hskip 1em plus 0.5em minus 0.4em\relax IEEE, 2013, pp. 829--834.

\bibitem{or2022hybrid}
B.~{O}r and I.~Klein, ``A {H}ybrid {M}odel and {L}earning-{B}ased {A}daptive {N}avigation {F}ilter,'' \emph{IEEE Transactions on Instrumentation and Measurement}, pp. 1--1, 2022.

\bibitem{wahlstrom2017smartphone}
J.~Wahlstr{\"o}m, I.~Skog, and P.~H{\"a}ndel, ``Smartphone-based vehicle telematics: A ten-year anniversary,'' \emph{IEEE Transactions on Intelligent Transportation Systems}, vol.~18, no.~10, pp. 2802--2825, 2017.

\bibitem{chan2019comprehensive}
T.~K. Chan, C.~S. Chin, H.~Chen, and X.~Zhong, ``A comprehensive review of driver behavior analysis utilizing smartphones,'' \emph{IEEE Transactions on Intelligent Transportation Systems}, vol.~21, no.~10, pp. 4444--4475, 2019.

\bibitem{mozaffari2020deep}
S.~Mozaffari, O.~Y. Al-Jarrah, M.~Dianati, P.~Jennings, and A.~Mouzakitis, ``Deep learning-based vehicle behavior prediction for autonomous driving applications: A review,'' \emph{IEEE Transactions on Intelligent Transportation Systems}, vol.~23, no.~1, pp. 33--47, 2020.

\bibitem{or2024carspeednet}
B.~Or, ``Carspeednet: A deep neural network-based car speed estimation from smartphone accelerometer,'' \emph{arXiv preprint arXiv:2401.07468}, 2024.

\bibitem{lecun2015deep}
Y.~LeCun, Y.~Bengio, and G.~Hinton, ``Deep learning,'' \emph{nature}, vol. 521, no. 7553, pp. 436--444, 2015.

\bibitem{bengio2017deep}
Y.~Bengio, I.~Goodfellow, and A.~Courville, \emph{Deep learning}.\hskip 1em plus 0.5em minus 0.4em\relax MIT press Cambridge, MA, USA, 2017, vol.~1.

\bibitem{yan2018ridi}
H.~Yan, Q.~Shan, and Y.~Furukawa, ``{RIDI}: Robust {IMU} double integration,'' in \emph{Proceedings of the European Conference on Computer Vision (ECCV)}, 2018, pp. 621--636.

\bibitem{brossard2019learning}
M.~Brossard and S.~Bonnabel, ``Learning wheel odometry and imu errors for localization,'' in \emph{2019 International Conference on Robotics and Automation (ICRA)}.\hskip 1em plus 0.5em minus 0.4em\relax IEEE, 2019, pp. 291--297.

\bibitem{or2022learning}
B.~Or and I.~Klein, ``Learning vehicle trajectory uncertainty,'' \emph{arXiv preprint arXiv:2206.04409}, 2022.

\bibitem{or2021kalman}
B.~Or, B.-Z. Bobrovsky, and I.~Klein, ``Kalman filtering with adaptive step size using a covariance-based criterion,'' \emph{IEEE Transactions on Instrumentation and Measurement}, vol.~70, pp. 1--10, 2021.

\bibitem{liu2021vehicle}
J.~Liu and G.~Guo, ``Vehicle localization during gps outages with extended kalman filter and deep learning,'' \emph{IEEE Transactions on Instrumentation and Measurement}, vol.~70, pp. 1--10, 2021.

\bibitem{freydin2022learning}
M.~Freydin and B.~Or, ``Learning car speed using inertial sensors for dead reckoning navigation,'' \emph{IEEE Sensors Letters}, vol.~6, no.~9, pp. 1--4, 2022.

\bibitem{freydin2022mountnet}
M.~Freydin, N.~Sfaradi, N.~Segol, A.~Eweida, and B.~Or, ``Mountnet: Learning an inertial sensor mounting angle with deep neural networks,'' \emph{arXiv preprint arXiv:2212.11120}, 2022.

\bibitem{shao2018indoor}
W.~Shao, H.~Luo, F.~Zhao, Y.~Ma, Z.~Zhao, and A.~Crivello, ``Indoor positioning based on fingerprint-image and deep learning,'' \emph{Ieee Access}, vol.~6, pp. 74\,699--74\,712, 2018.

\bibitem{or2023system}
B.~Or, M.~Freydin, and G.~Ben-haim, ``System and method for estimating a location of a vehicle using inertial sensors,'' Jan.~3 2023, uS Patent 11,543,245.

\bibitem{or2023patent1}
B.~Or, ``System and method for providing localization using inertial sensors,'' Aug.~15 2023, uS Patent 11,725,945 B2.

\bibitem{tianyi2019}
T.~Li, M.~Yang, H.~Li, L.~Deng, and C.~Wang, ``A terrain-based vehicle localization approach robust to braking,'' \emph{IEEE Transactions on Intelligent Transportation Systems}, vol.~20, no.~8, pp. 2923--2932, 2019.

\bibitem{adam2008}
A.~J. Dean, R.~D. Martini, and S.~N. Brennan, ``Terrain-based road vehicle localization using particle filters,'' in \emph{2008 American Control Conference}, 2008, pp. 236--241.

\bibitem{laftchiev2015}
E.~I. Laftchiev, C.~M. Lagoa, and S.~N. Brennan, ``Vehicle localization using in-vehicle pitch data and dynamical models,'' \emph{IEEE Transactions on Intelligent Transportation Systems}, vol.~16, no.~1, pp. 206--220, 2015.

\bibitem{emulapalli2011}
P.~K. Vemulapalli, A.~J. Dean, and S.~N. Brennan, ``Pitch based vehicle localization using time series subsequence matching with multi-scale extrema features,'' in \emph{Proceedings of the 2011 American Control Conference}, 2011, pp. 2405--2410.

\bibitem{eweida2023surface}
A.~Eweida, N.~Segol, M.~Freydin, N.~Sfaradi, and B.~Or, ``Surface recognition for e-scooter using smartphone imu sensor,'' in \emph{2023 8th International Conference on Signal and Image Processing (ICSIP)}.\hskip 1em plus 0.5em minus 0.4em\relax IEEE, 2023, pp. 1107--1111.

\bibitem{barak2024system}
O.~Barak, ``System and method for estimating a velocity of a vehicle using inertial sensors,'' Jan.~2 2024, uS Patent 11,859,978.

\bibitem{barak2023system}
O.~Barak, M.~Freydin, and G.~Ben-Haim, ``System and method for estimating a location of a vehicle using inertial sensors,'' Jan.~3 2023, uS Patent 11,543,245.

\bibitem{ismail2019deep}
H.~Ismail~Fawaz, G.~Forestier, J.~Weber, L.~Idoumghar, and P.-A. Muller, ``Deep learning for time series classification: a review,'' \emph{Data mining and knowledge discovery}, vol.~33, no.~4, pp. 917--963, 2019.

\bibitem{zhang2012ensemble}
C.~Zhang and Y.~Ma, \emph{Ensemble machine learning: methods and applications}.\hskip 1em plus 0.5em minus 0.4em\relax Springer, 2012.

\bibitem{rajamani2011vehicle}
R.~Rajamani, \emph{Vehicle dynamics and control}.\hskip 1em plus 0.5em minus 0.4em\relax Springer Science \& Business Media, 2011.

\bibitem{kingma2014adam}
D.~P. Kingma and J.~Ba, ``Adam: A method for stochastic optimization,'' \emph{arXiv preprint arXiv:1412.6980}, 2014.

\bibitem{bennett2010openstreetmap}
J.~Bennett, \emph{OpenStreetMap}.\hskip 1em plus 0.5em minus 0.4em\relax Packt Publishing Ltd, 2010.

\bibitem{staacks2018advanced}
S.~Staacks, S.~H{\"u}tz, H.~Heinke, and C.~Stampfer, ``Advanced tools for smartphone-based experiments: phyphox,'' \emph{Physics education}, vol.~53, no.~4, p. 045009, 2018.

\end{thebibliography}

\begin{IEEEbiography}[{\includegraphics[width=1in,height=1.25in,clip,keepaspectratio]{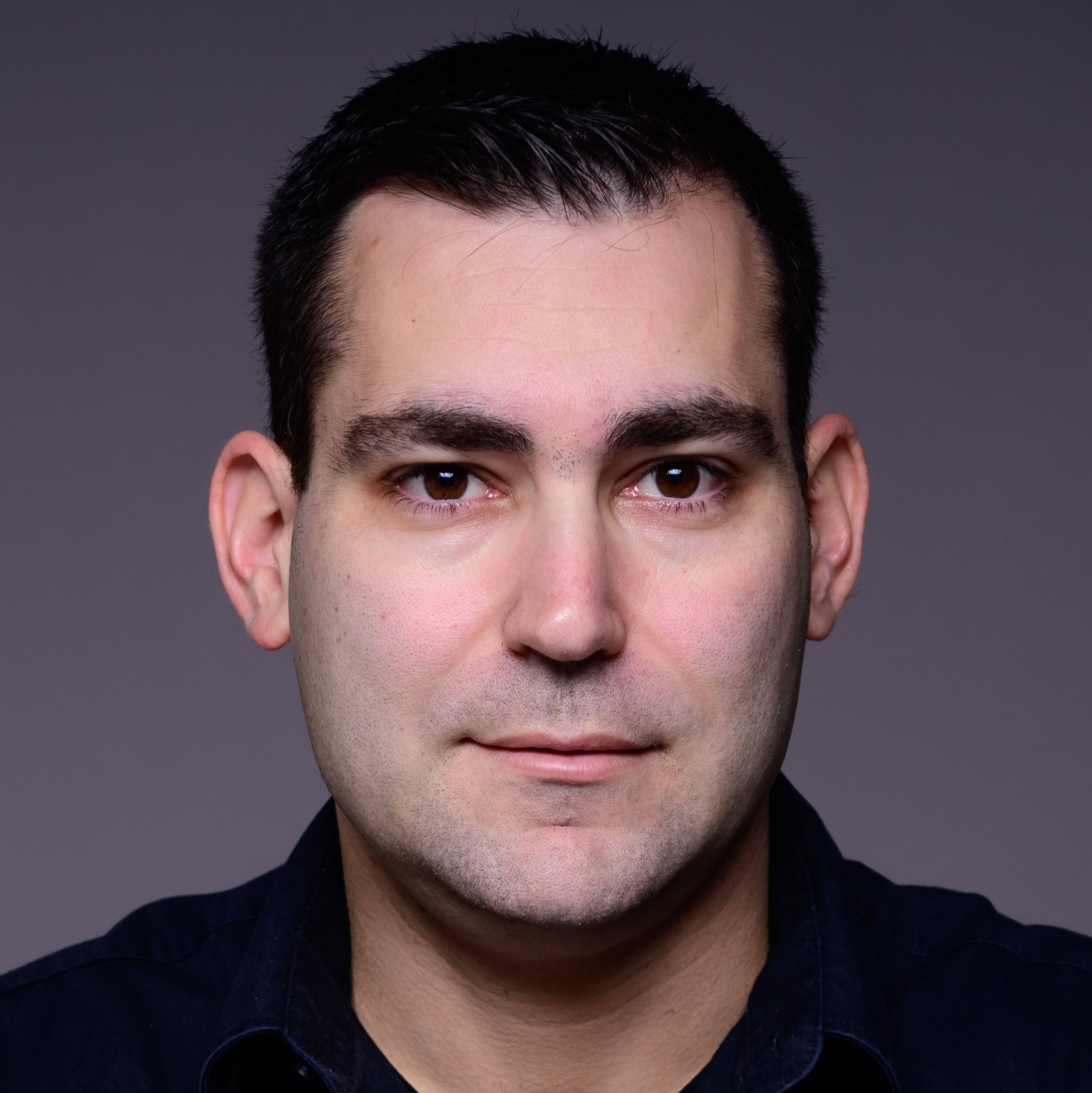}}]{Barak Or} (Member, IEEE) received a B.Sc. degree in aerospace engineering (2016), a B.A. degree (cum laude) in economics and management (2016), and an M.Sc. degree in aerospace engineering (2018) from the Technion–Israel Institute of Technology. He graduated with a Ph.D. degree from the University of Haifa, Haifa (2022).
He founded ALMA Technologies Ltd. (2021). His research interests include navigation, deep learning, sensor fusion, and estimation theory.
\end{IEEEbiography}

\begin{IEEEbiography}[{\includegraphics[width=1in,height=1.25in,clip,keepaspectratio]{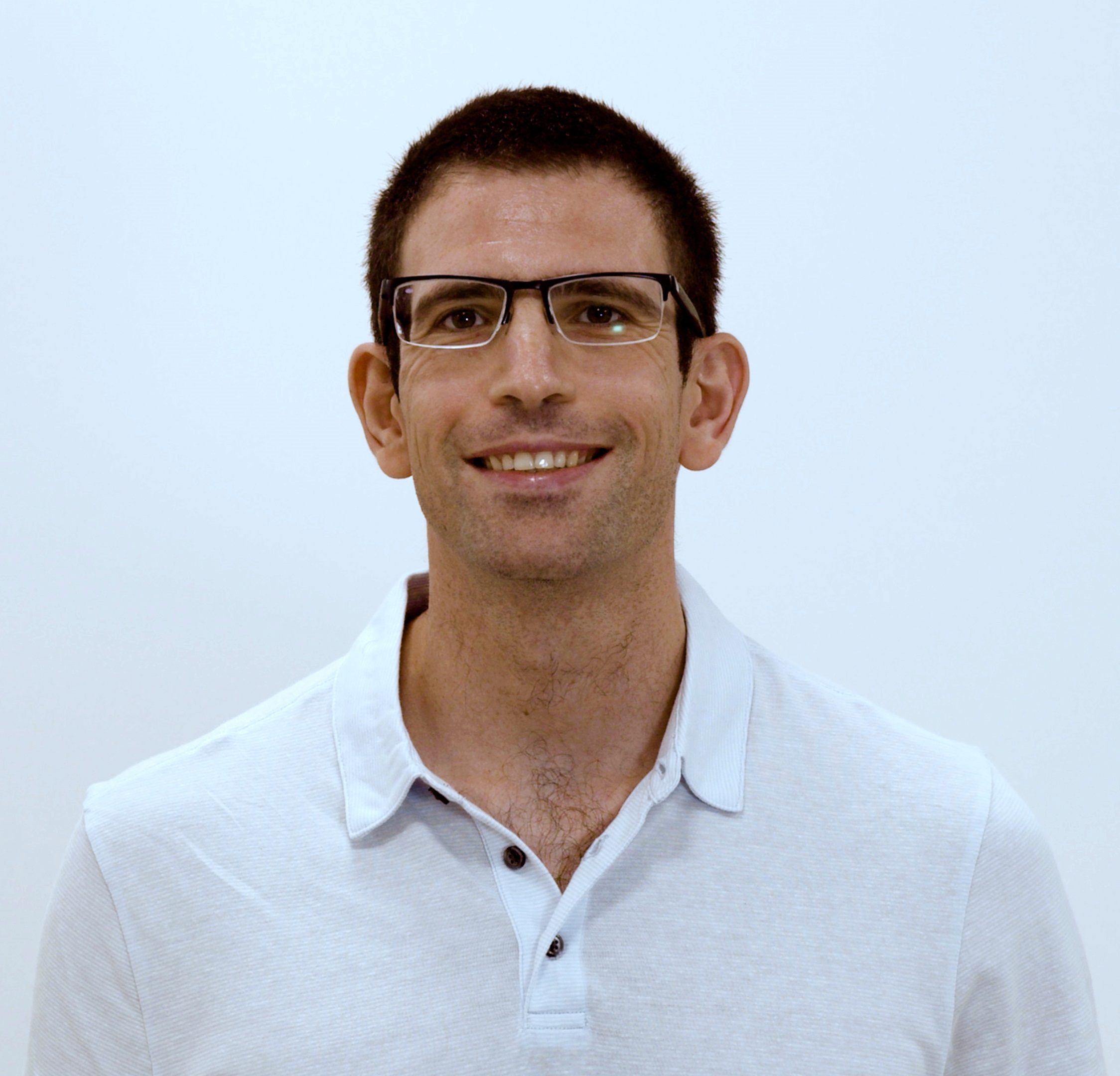}}]{Nimrod Segol}  received his B.Sc. (2015,  Cum Laude) and M.Sc. (2018) degrees in Mathematics from the Technion – Israel Institute of Technology. He is a senior algorithm engineer at ALMA Technologies Ltd. and his research interests include deep learning, ML, and statistics.
\end{IEEEbiography}

\begin{IEEEbiography}[{\includegraphics[width=1in,height=1.25in,clip,keepaspectratio]{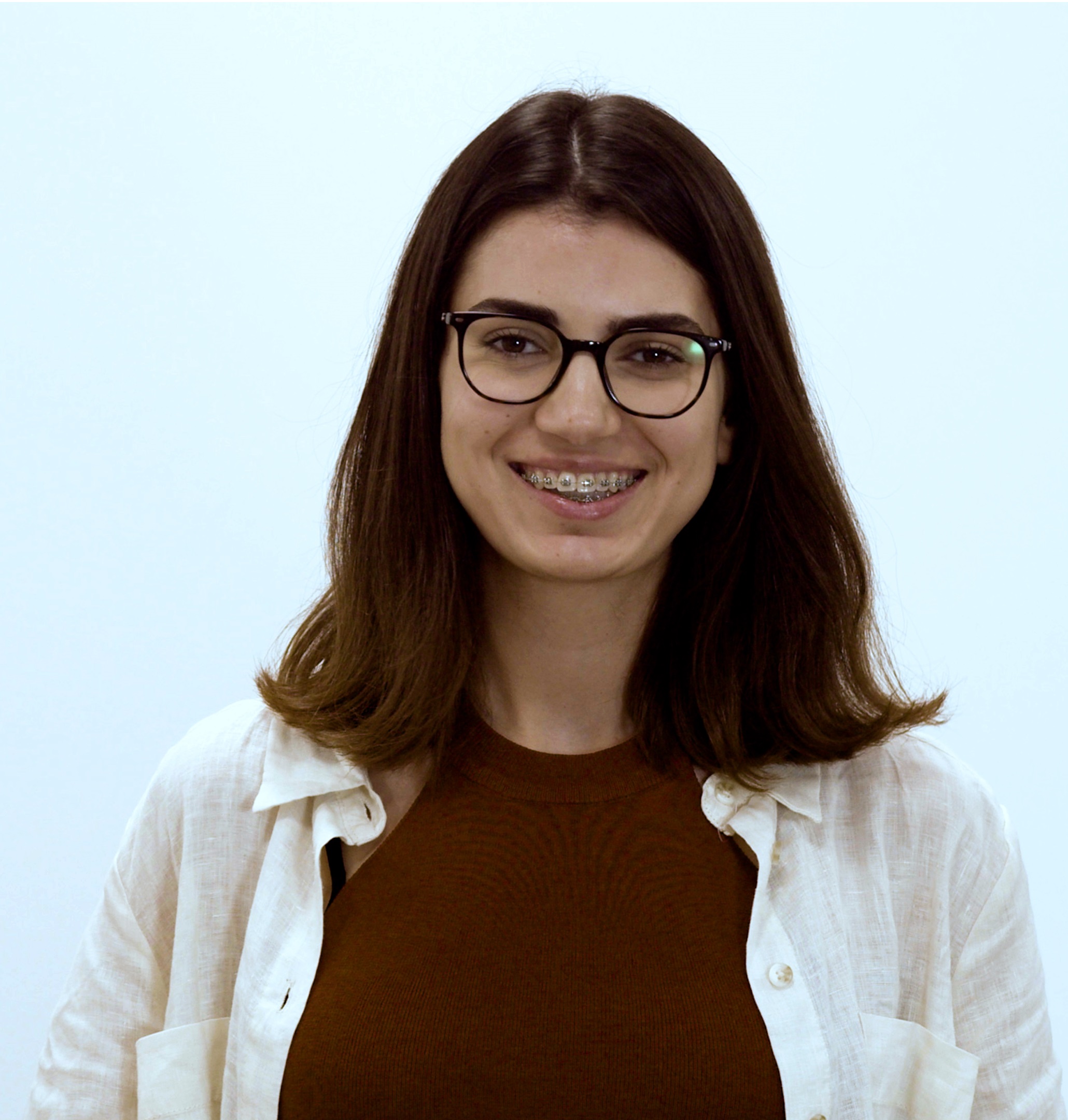}}]{Areej Eweida} received her B.Sc. (2022) degree in Aerospace Engineering from the Technion – Israel Institute of Technology. She is an algorithm engineer at ALMA Technologies Ltd. She is starting an M.Sc. Her research interests include estimation theory, signal processing, and deep learning.
\end{IEEEbiography}

\begin{IEEEbiography}[{\includegraphics[width=1in,height=1.25in,clip,keepaspectratio]{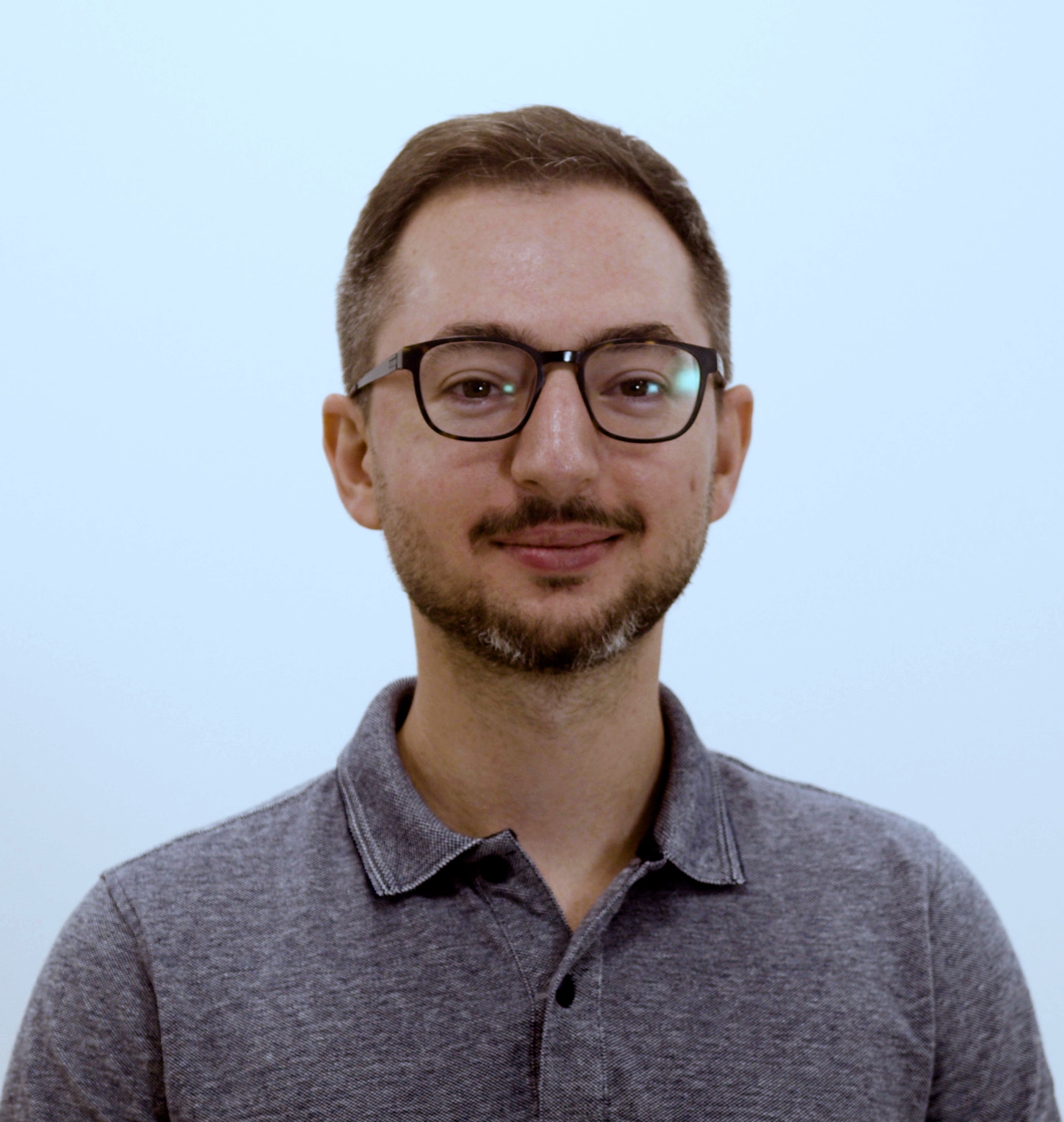}}]{Maxim Freydin} (Member, IEEE) received his B.Sc. (2017, Summa Cum Laude) and Ph.D. (2021) degrees in Aerospace Engineering from the Technion – Israel Institute of Technology. He is the VP of R\&D at ALMA Technologies Ltd. and his research interests include navigation, signal processing, deep learning, and computational fluid-structure interaction.
\end{IEEEbiography}

\end{document}